%% file: main.tex
\documentclass{article}

\makeatletter

\IfFileExists{template/iclr2027_conference.sty}{}{%
  
\edef\input@path{{template/}}}
\makeatother

\usepackage{template/iclr2027_conference}
\usepackage{times}
\input{template/math_commands.tex}

\usepackage{graphicx}
\usepackage{hyperref}
\usepackage{etoolbox}
\usepackage{url}

\title{Soundwich: Video Generation with Layered and Controllable Audio}
\newcommand{\methodname}{Soundwich\xspace}
\author{\textbf{Zhuo Ning}$^{1}$\quad \textbf{AmirHossein Naghi Razlighi}$^{2,3}$\quad \textbf{Sagi Polaczek}$^{4}$\\
\textbf{Daniel Cohen-Or}$^{4}$\quad \textbf{Ali Mahdavi-Amiri}$^{1}$\\
{\normalfont $^{1}$Simon Fraser University\quad $^{2}$University of Cyprus}\\
{\normalfont $^{3}$CYENS Centre of Excellence\quad $^{4}$Tel Aviv University}}

\iclrfinalcopy

\include{sections/commands}

\input{sections/study_results}
\AtEndEnvironment{abstract}{ Code is available at \url{https://github.com/CodyNing/Soundwich}.}
\begin{document}

\maketitle
\lhead{Preprint.}

\begin{center}
    \centering
    \captionsetup{type=figure}
    \includegraphics[width=\linewidth]{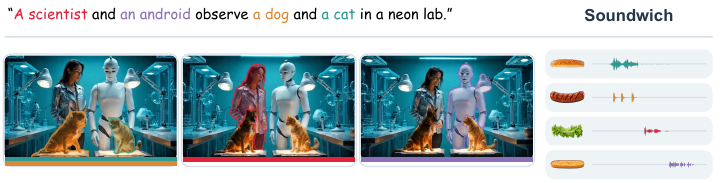}
    \captionof{figure}{\textbf{Soundwich generates layered audio for a shared video.} Given a scene with $N$ sound sources, our method generates a shared video together with $N$ separate and editable audio stems, one for each source. Like the ingredients of a sandwich, the sounds form distinct layers that remain synchronized with the same scene and can be independently controlled, mixed, muted, or replaced.}
    \label{fig:teaser}
\end{center}

\input{sections/abstract}
\input{sections/intro2}
\input{sections/related_work}
\input{sections/method_compressed}
\input{sections/experiments}
\input{sections/conclusion}

\clearpage
\bibliography{references}
\bibliographystyle{template/iclr2027_conference}

\clearpage
\appendix

\input{sections/appendix_supplementary}
\input{sections/appendix_method_details}
\input{sections/appendix_benchmark}
\input{sections/appendix_evaluation_protocols}
\input{sections/appendix_additional_results}
\input{sections/appendix_applications}

\end{document}

%% file: template/math_commands.tex
\usepackage{amsmath,amsfonts,bm}

\def\eqref#1{equation~\ref{#1}}

\def\1{\bm{1}}

\DeclareMathAlphabet{\mathsfit}{\encodingdefault}{\sfdefault}{m}{sl}
\SetMathAlphabet{\mathsfit}{bold}{\encodingdefault}{\sfdefault}{bx}{n}

%% file: sections/commands.tex
\usepackage{microtype}

\renewcommand{\paragraph}[1]{\vspace{.5em}\noindent\textbf{#1}}

\usepackage{color}
\usepackage{graphicx}
\usepackage{caption}
\usepackage{subcaption}
\usepackage[leftcaption,raggedright]{sidecap}
\usepackage{tabularx}
\usepackage{wrapfig}

\usepackage{amssymb}
\usepackage{amsmath}
\usepackage[utf8]{inputenc}
\usepackage{pifont}
\usepackage{makecell}
\usepackage{float}
\usepackage{array}
\usepackage{booktabs}
\usepackage{longtable}
\usepackage{colortbl}
\usepackage{multirow}
\usepackage{amsmath}
\usepackage{dblfloatfix}
\usepackage{xspace}

\definecolor{title_purple}{rgb}{0.65,0.1,0.65}
\definecolor{title_col_1}{rgb}{0.827703, 0.147925, 0.127676}
\definecolor{title_col_2}{rgb}{0.842658, 0.229482, 0.179246}
\definecolor{title_col_3}{rgb}{0.856618, 0.293874, 0.225622}
\definecolor{title_col_4}{rgb}{0.869563, 0.350689, 0.269494}
\definecolor{title_col_G}{rgb}{0.5, 0.5, 0.5}

\definecolor{darkred}{rgb}{0.7,0.1,0.1}
\definecolor{darkgreen}{rgb}{0.1,0.6,0.1}
\definecolor{green}{rgb}{0, 0.5, 0}
\definecolor{orange}{rgb}{0.8, 0.6, 0.2}
\definecolor{red}{rgb}{1.0, 0.0, 0.0}
\definecolor{teal}{rgb}{0.0, 0.4, 0.4}
\definecolor{purple}{rgb}{0.65,0.0,0.65}
\definecolor{saffron}{rgb}{0.95,0.75,0.2}
\definecolor{turquoise}{rgb}{0.0,0.5,0.5}
\definecolor{black}{rgb}{0.0, 0.0, 0.0}
\definecolor{gray}{rgb}{0.5, 0.5, 0.5}

\definecolor{titleblue}{RGB}{70,80,210}

%% file: sections/study_results.tex
\newcommand{\MainResponses}{18}
\newcommand{\AblationResponses}{15}
\newcommand{\MainTrials}{270}
\newcommand{\AblationTrials}{150}
\newcommand{\MainFullNat}{1.16}
\newcommand{\MainLTXNat}{2.22}
\newcommand{\MainNaiveNat}{2.62}
\newcommand{\MainFullTime}{1.06}
\newcommand{\MainLTXTime}{2.25}
\newcommand{\MainNaiveTime}{2.69}
\newcommand{\MainFullSep}{74.4}
\newcommand{\MainNaiveSep}{25.6}
\newcommand{\MainFullNatFirst}{87.0}
\newcommand{\MainFullTimeFirst}{95.2}
\newcommand{\FullContext}{1.27}
\newcommand{\NoBroadcastContext}{2.04}
\newcommand{\StageOneContext}{2.69}
\newcommand{\FullQuality}{8.19}
\newcommand{\NoBroadcastQuality}{6.51}
\newcommand{\StageOneQuality}{4.66}
\newcommand{\FullOverall}{1.27}
\newcommand{\NoBroadcastOverall}{2.03}
\newcommand{\StageOneOverall}{2.70}
\newcommand{\FullContextFirst}{76.7}
\newcommand{\NoBroadcastContextFirst}{17.3}
\newcommand{\FullOverallFirst}{76.0}
\newcommand{\NoBroadcastOverallFirst}{18.7}
\newcommand{\FullActivation}{96.0}
\newcommand{\NoRoutingActivation}{40.7}
\newcommand{\FullOwnership}{95.3}
\newcommand{\NoRoutingOwnership}{34.7}
\newcommand{\MainSepFullWins}{159}
\newcommand{\MainSepNaiveWins}{27}
\newcommand{\MainSepTies}{84}
\newcommand{\MainNatWins}{235}
\newcommand{\MainTimeWins}{257}

\newcommand{\StudioUnroutedYes}{3}

%% file: sections/abstract.tex
\begin{abstract}
Recent joint audio-video generative models can synthesize realistic videos with synchronized sound, but typically generate audio as a single mixed track. This limits source-level control and differs from practical audiovisual workflows, where speech, music, sound effects, and ambient sounds are represented as separate editable tracks. We introduce \methodname, a training-free framework that transforms a frozen joint audio-video flow-matching model into a generator of multiple synchronized, independently editable audio stems coupled to a shared video. \methodname generates separate audio stems with explicit control over their temporal activity. To keep separately generated sounds coherent, we introduce a shared scene representation that communicates global audiovisual context across stems while preserving their source-level separation. We further route cross-modal interactions between each audio stem and its corresponding visual source, improving audiovisual consistency. The resulting stems remain synchronized with the video and can be independently retimed, muted, replaced, or remixed.
Experiments and human evaluations show improved temporal control, source separation, and naturalness, while enabling flexible source-level editing within coherent audiovisual generation.
\end{abstract}

%% file: sections/intro2.tex
\section{Introduction}

Recent advances in generative models have made it possible to synthesize increasingly realistic videos together with their corresponding audio. Joint audio-video generation is particularly appealing because sound and visual content are generated within a shared process, allowing speech, object interactions, music, and ambient sounds to emerge in synchronization. Yet, despite impressive progress in generation quality, these models largely treat audio as a single monolithic output. This is fundamentally different from how audiovisual content is created and edited in practice, where speech, music, sound effects, and ambience are typically represented as separate layers that can be independently manipulated.

Our goal is to bring this \emph{layered} representation to joint audio-video generation. Given a scene with multiple sound sources, we aim to generate a single video together with synchronized audio stems, each corresponding to a particular source and independently \emph{controllable}. For example, a scene with two people speaking over background music would produce separate stems for the two speakers and the music, all synchronized with the same generated video. Beyond separating these sources, the layered representation provides natural handles for controlling them: a source can be specified by its semantic \emph{content} and, when desired, by when it should be \emph{active} (e.g., sec. 3--5).
Notably, temporal control is optional: Soundwich can separate sources even when their activity fully overlaps (see Section~\ref{sec:ablation-results}).
The resulting stems can then be retimed, muted, replaced, or remixed independently.

Achieving such a layered representation with a pretrained joint audio-video
model is not straightforward. The model is designed to evolve a single audio
trajectory jointly with a video trajectory, ultimately producing one mixed
soundtrack. A natural way to introduce layers is to expand the audio trajectory
into a batch of source-specific trajectories, each with its own conditioning,
while retaining a single shared video trajectory. However, this construction
introduces a fundamental challenge. The audio stems need to share information
through the common video to remain coherent parts of the same scene, while this
shared representation also creates a path for interference: sound from one stem
may leak into another or become associated with the wrong visual entity.

We introduce \emph{Soundwich}, named after its layered representation of sound, a training-free framework that addresses this challenge. Soundwich generates multiple source-specific audio trajectories in parallel while maintaining a single shared video trajectory, as illustrated in Figure~\ref{fig:teaser}. It carefully controls how information is shared across these trajectories, allowing the stems to benefit from common scene context while limiting cross-source leakage.
Soundwich organizes this process around two complementary roles: Stem Formation establishes each source stem's identity, content, and timing, while Scene Integration coordinates the stems through shared scene context and associates them with their corresponding visual sources. On top of this layered generation, Soundwich provides additional control handles over individual sources, including their semantic content and, when desired, their temporal activity.

Our main contribution is a training-free framework for layered audio-video generation from a pretrained joint generator. Soundwich enables (i) the joint generation of multiple source-specific audio stems with a shared video, (ii) source-level semantic and temporal control, and (iii) coherent interaction between the stems and their corresponding visual sources. Together, these capabilities turn a monolithic generated soundtrack into a layered and editable representation, bringing joint audio-video generation closer to practical audiovisual creation workflows.

%% file: sections/related_work.tex
\section{Related Work}
\label{sec:related-work}

\paragraph{Joint audio-video and multi-stem generation.}
Approaches to joint audio-video generation include coupled modality-specific denoisers~\cite{ruan2023mmdiffusion}, inference-time alignment of separately pretrained generators~\cite{xing2024seeing}, dual-stream transformer architectures such as SyncFlow, Ovi, UniAVGen, and LTX-2~\cite{liu2024syncflow,low2025ovi,zhang2026uniavgen,ltx_2}, and single-stream transformers such as MiniMax H3~\cite{minimax2026h3}.
These joint audiovisual priors have also been adapted for video dubbing~\cite{justdubit}, instruction-guided audio-video editing~\cite{instructav2av}, and joint appearance and voice personalization~\cite{idlora}.
Such systems generate a single mixed soundtrack, rather than a set of isolated audio stems that can be independently retimed, remixed, or removed. Audio-only work has explored structured multi-source generation more directly. Generalized Multi-Source Inference constructs a variable set of coherent music sources at inference time from text-conditioned diffusion models~\cite{gmsi}, while MusicGen-Stem, MGE-LDM, and Stemphonic learn multi-track generation, editing, imputation, or source extraction from music data~\cite{musicgen,chae2025mgeldm,wu2026stemphonic}. These methods establish the value of joint stem modeling, but remain audio-only and predominantly music-specific. Step-by-Step V2A generates complementary tracks for a fixed video using a fine-tuned guidance model, conditioning on earlier audio for acoustic context while discouraging duplication across tracks~\cite{hayakawa2026stepbystep}. \methodname instead adapts a frozen joint audio-video generator to produce speech, music, and general sounds as separate stems coupled to one shared video.

\paragraph{Guidance and temporal control.}
Independent source conditions do not by themselves guarantee source identity or event timing. Classifier-free and compositional guidance provide general mechanisms for strengthening and combining conditions~\cite{ho2022cfg,liu2022composable}, while temporally controllable audio models incorporate timestamp conditions, structured scripts, or learned temporal controllers~\cite{picoaudio,foleycrafter,li2026foleydirector,li2026echofoley,jiang2026controlaudio}. FreeAudio instead manipulates attention and composes latent windows to obtain training-free timing control in a text-to-audio model~\cite{jiang2025freeaudio}. Activation steering provides another inference-time control mechanism: CASteer derives directions from cross-attention activations and injects them to add or suppress image concepts without retraining~\cite{gaintseva2026casteer}. Unlike these settings, \methodname must coordinate the timing of multiple isolated sources with a jointly evolving video while retaining every source as an editable output.

\paragraph{Audiovisual source grounding and entity control.}
Audiovisual grounding and separation methods associate sounding regions or visual objects with components of an observed audio mixture~\cite{sound_of_pixels,coseperation,cyclic}. SAM Audio provides general source separation with text, visual, and temporal prompts~\cite{shi2025samaudio}. Video-to-audio systems also introduce source-level generation control: DualDub uses separate decoders for speech and background audio~\cite{tian2025dualdub}, SelVA selects a text-specified source from a multi-object video~\cite{lee2026selva}, and SAGANet conditions sound synthesis on a selected object's segmentation mask~\cite{viertola2026saganet}. These video-to-audio methods operate on a fixed input video.
In joint audio-video generation, attention steering has been used to reduce cross-modal semantic leakage and improve source grounding~\cite{attention_triangle}. To ground individual output stems, \methodname segments entities in its generated video and restricts the stem-video attention of each output stem to its corresponding entity during audiovisual refinement.

\paragraph{Shared attention across diffusion trajectories.}
Shared-attention methods couple diffusion trajectories to preserve style, appearance, or source content across generated images~\cite{hertz2024stylealigned,cao2023masactrl}. Untwisting RoPE further analyzes how positional encoding can cause shared-attention queries to copy spatially aligned reference content~\cite{mikaeili2026untwisting}. \methodname uses cross-trajectory attention for a different purpose: an asymmetric gather-broadcast topology lets a persistent scene stem collect context from the active source stems and return shared acoustic information to them. This coordination preserves common scene context while keeping the decoded stems separate and independently editable.

%% file: sections/method_compressed.tex
\vspace{-5pt}
\section{Method}
\label{sec:method}
\vspace{-5pt}

\begin{figure}[t]
    \centering
    \includegraphics[width=\linewidth]{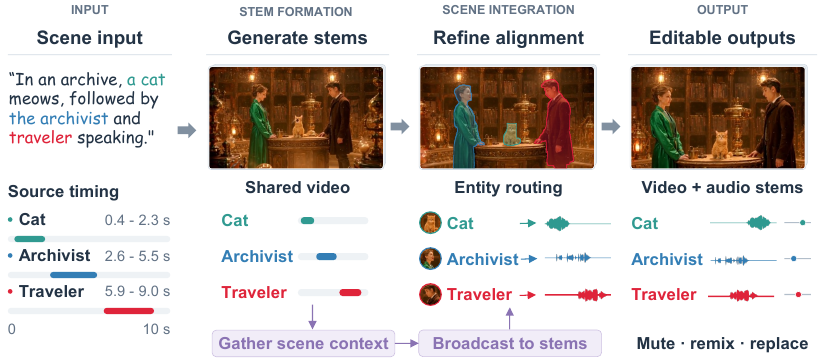}
    \caption{\textbf{Overview of \methodname}.
    Stem Formation expands the audio
    trajectory into $N$ editable source stems and one internal scene stem while
    retaining a shared video trajectory. Per-stem CFG and cached carrier replay
    isolate and schedule the sources, while the scene stem gathers shared
    context.
    Scene Integration uses entity-aware cross-modal attention to associate each
    stem with its corresponding visual source and broadcasts scene context back
    to the stems. The resulting $N$ synchronized audio stems can be independently
    retimed, muted, replaced, or remixed.}
    \label{fig:method-overview}
\end{figure}

A conventional joint audio--video generator evolves a single audio trajectory
together with a video trajectory, producing one video with a mixed soundtrack.
\methodname instead expands the audio trajectory into $N$ source-specific
trajectories while retaining a shared video trajectory. Each trajectory produces
a separately editable audio track, which we call a \emph{source stem}.

The layered representation exposes several controls over each source. We denote
the visual prompt by $p^v$. For source $i$, $p_i^+$ describes the desired sound,
while $p_i^-$ specifies competing sounds to exclude. Optionally,
$\mathcal W_i$ specifies the intervals in which the source should be active;
when $\mathcal W_i$ is empty, its timing is left unconstrained. For visible
sound-producing entities, $q_i$ describes the corresponding entity and is used
to obtain its video segmentation mask; $q_i$ is omitted for audio-only sources
such as background noise or music. Different sources may have overlapping
activity intervals, allowing simultaneous sounds while keeping them on separate
editable stems. Together,
$\mathcal S=\{(p_i^+,p_i^-,\mathcal W_i,q_i)\}_{i=1}^N$
conditions a video $x^v$ and audio stems $\{x_i^a\}_{i=1}^N$.

\methodname operates on a fixed, pretrained joint audio--video flow-matching
transformer with frozen parameters $\theta$~\cite{lipman2023flowmatching}.
At noise level $\sigma$, the model takes noisy audio and video states
$(z_\sigma^a,z_\sigma^v)$ and predicts a velocity field. For modality
$\mu\in\{a,v\}$ and text condition $y$, we denote this by
$v_\theta^{\mu,y}$ and obtain the clean estimate as:
\begin{equation}
D_\theta^{\mu,y}(\sigma)
=z_\sigma^\mu-\sigma v_\theta^{\mu,y}
(z_\sigma^a,z_\sigma^v,\sigma).
\label{eq:flow-prediction}
\end{equation}
Classifier-free guidance (CFG) combines positive and negative conditional
predictions. The transformer uses separate audio and video self-attention,
together with audio-to-video (A2V) and video-to-audio (V2A) attention.
\methodname modifies the intermediate audio states and attention connections
without changing the pretrained model parameters.

Our main technical challenge is to coordinate the parallel audio trajectories
through the shared video while preventing information from leaking across
sources.
We address this via Stem Formation and Scene Integration. Stem Formation establishes each source stem’s identity, content, and timing while collecting shared context in an internal scene stem. Scene Integration combines the stems into a coherent audiovisual scene by broadcasting shared scene context and using entity-aware cross-modal attention to associate each stem with its visual source. Figure~\ref{fig:method-overview} summarizes the
method.

\subsection{Stem Formation}
\label{sec:stage_1}

\noindent{\textbf{Source Stems and Per-Stem Guidance.}} To produce multiple audio stems, we batch-expand the audio latent into $N$ source-specific latent trajectories $\{z_{\sigma,i}^a\}_{i=1}^{N}$, which we call source stems, while retaining a shared video latent. Each source stem has its own latent trajectory and audio condition while sharing the evolving video. We add an internal scene stem $z_{\sigma,\mathrm{sc}}^a$ to collect information across these separately represented sources for subsequent refinement.

When multiple sounds are complex and strongly entangled, separating them becomes challenging, and a single positive prompt is often not sufficient. For example, a joint generator may treat concurrent piano and violin as one musical event and reproduce both instruments in either audio stream.
We therefore apply standard classifier-free guidance (CFG) (~\cite{ho2022cfg}) independently to every stem. The positive prompt $p_i^+$ specifies the target sound, and the negative prompt $p_i^-$ names competing sources to exclude. CFG steers the stem toward its requested source and away from competing content. The scene stem receives a joint scene condition because it must collect context from multiple sources. More details are provided in Appendix~\ref{app:source-guidance}.

\noindent{\textbf{Timeline Control with Cached Carriers.}} Per-stem guidance specifies what to generate; a separate temporal signal determines when to generate it. Our idea is to obtain reusable examples of sound and quiet from the model itself. We generate a reference containing the target sound, such as a meowing cat, and a second reference containing silence. Within the transformer, audio self-attention produces a feature vector at each audio-token position. We average selected reference vectors into a single vector, called a \emph{carrier}, that can be added or blended into the corresponding features of a new generation.

\begin{figure}[tbp]
    \centering
    \includegraphics[width=0.95\linewidth]{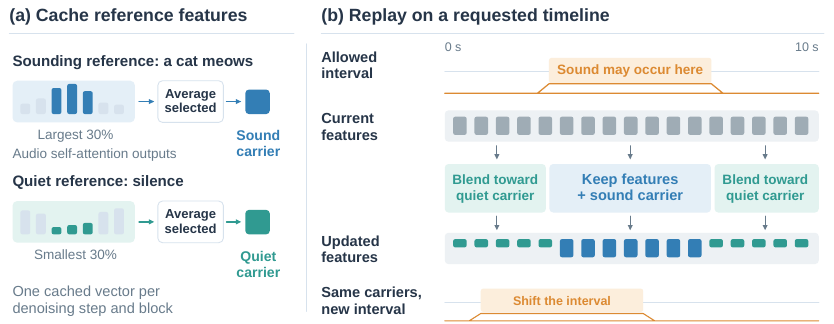}
    \caption{\textbf{Cache sound and quiet features, then replay them at new times.} \textbf{(a)} A sounding reference and a silent reference provide audio self-attention outputs. Averaging selected high- or low-magnitude vectors produces one carrier per step and block. \textbf{(b)} The activation carrier is added to the target's current self-attention output within its requested interval; outside, the output is attenuated and blended toward the quiet carrier. Changing the requested interval changes where the cached features
    are replayed, so the same reference generation can guide a sound
    at different times. Reference-bar heights indicate feature magnitude; the shaded bands show temporal guidance.}
    \label{fig:carrier-timeline}
\end{figure}

For each denoising step $n$ and block $\ell$, we average high-magnitude vectors from the sounding reference into an \emph{activation carrier} $c^+_{i,n,\ell}$ and low-magnitude vectors from the silent reference into a \emph{suppression carrier} $s_{n,\ell}$. Magnitude is measured by root-mean-square (RMS) across feature channels. Averaging removes the explicit sequence of reference positions, so the cached vector can be repeated at newly requested times (Figure~\ref{fig:carrier-timeline}). Each source uses a matching activation reference, while sources share the quiet reference.

During target generation, we add the activation carrier to the model's current self-attention output inside the allowed intervals. Outside them, we attenuate that output and blend it toward the quiet carrier. This second path controls unwanted off-window sound: at the same suppression strengths, a cached quiet target reduces off-window activity more effectively than a zero target (Appendix~\ref{app:zero-vs-quiet}). The resulting features continue through the pretrained transformer to generate the source audio.

Let $h_i(\tau)$ be the target stem's current self-attention output at audio time $\tau$. A soft activity gate $m_i(\tau)$ is one inside $\mathcal W_i$ and tapers linearly to zero over a short boundary region; multiple intervals use the maximum of their gates. The replay paths are summarized by
\begin{equation}
\begin{aligned}
\widetilde h_i(\tau)={}&
\underbrace{\bigl[1-\gamma_s(1-m_i(\tau))\bigr]h_i(\tau)}_{\text{retained audio features}}
+\underbrace{\gamma_s(1-m_i(\tau))s_{n,\ell}}_{\text{outside suppression}}\\[-2pt]
&+\underbrace{\lambda_n m_i(\tau)\rho_{i,n,\ell}c^+_{i,n,\ell}}_{\text{inside activation}}.
\end{aligned}
\label{eq:carrier-replay}
\end{equation}
Here $\rho_{i,n,\ell}$ scales activation to the receiving stem's RMS, $\lambda_n$ sets its denoising-dependent strength, and $\gamma_s$ summarizes off-window suppression. We use a short activation boost ($0.40$ for the first 10\% of denoising), followed by a smaller strength ($0.08$) as the sound develops. The quiet carrier keeps its recorded low RMS, and $0.20$-second boundary tapering makes the transition gradual. Appendix~\ref{app:cached-carriers} gives the reference selection, parameter settings, and exact factorization of this schematic update.

\subsection{Scene Integration}
\label{sec:entity-refinement}

\noindent\textbf{Scene-Based Cross-Stem Coordination.} Per-stem guidance and carrier replay help each stem generate the requested source at the correct time. However, keeping the stems fully independent can make a conversation or performance sound like a collection of unrelated recordings. Direct communication between stems can restore this shared context, but it can also cause sounds to leak from one stem to another, as shown in Appendix~\ref{app:direct-sharing}. We therefore use an internal scene stem as a shared information buffer. In Stem Formation, the scene stem gathers information from the active source stems, capturing their timing and interactions while keeping the source stems separate. In Scene Integration, the information flow is reversed: each source stem reads from the scene stem during its active interval to recover the shared context needed for more natural interactions. The scene stem is discarded after generation, leaving only the separate editable source stems.

For the attention operations below, $\mathcal E$ is the set of sources selected to share scene context; $Q$, $K$, and $V$ are the pretrained query, key, and value projections. $\operatorname{Attn}$ denotes scaled dot-product attention, and $[\,;\,]$ concatenates token sequences with aligned temporal positions.
During Stem Formation, the scene stem gathers timing and shared context without direct communication between source stems. Let $u_i(\tau)$ and $u_{\mathrm{sc}}(\tau)$ denote the source and scene states. We combine the active source states as $r(\tau)=\sum_{i\in\mathcal E}m_i(\tau)u_i(\tau)$ and let the scene attend to this representation:
\begin{equation}
g_{\mathrm{sc}}=\operatorname{Attn}\!\left(
Q(u_{\mathrm{sc}}),[K(u_{\mathrm{sc}});K(r)],[V(u_{\mathrm{sc}});V(r)]\right).
\label{eq:scene-gather}
\end{equation}
This captures the timing and coarse structure of the sources while keeping self-attention independent.

In Scene Integration, the information flow is reversed. Each source attends to itself and saved scene:
\begin{equation}
b_i=\operatorname{Attn}\!\left(
Q(u_i),[K(u_i);K(u_{\mathrm{sc}})],[V(u_i);V(u_{\mathrm{sc}})]\right).
\label{eq:scene-broadcast}
\end{equation}
This update is applied only during the source's active interval, allowing each stem to recover shared scene context without directly attending to other stems. The scene stem is discarded after generation.
Figure~\ref{fig:gather-broadcast} in Appendix~\ref{app:scene-sharing} illustrates both directions.

\begin{figure}[!tb]
    \centering
    \includegraphics[width=0.88\linewidth]{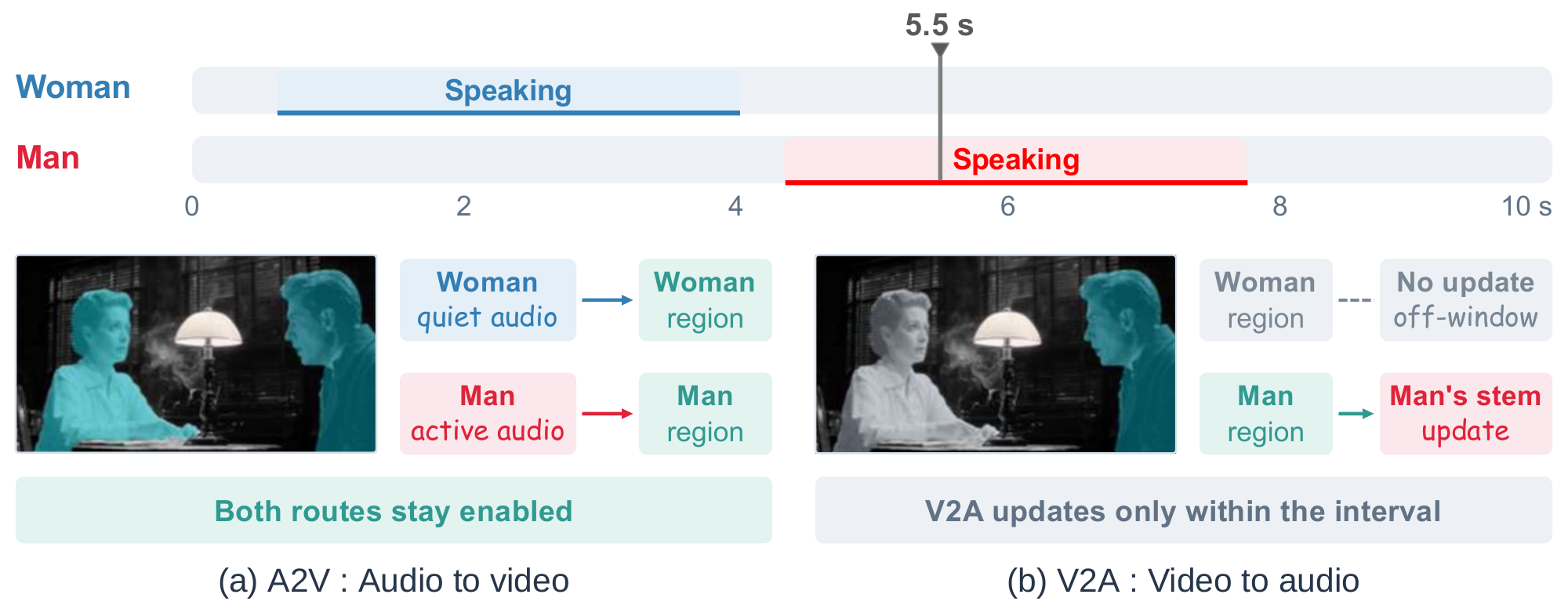}
\caption{\textbf{Mask-guided routing at the marked time: the woman is quiet and the man is speaking.} \textbf{(a)} A2V links each entity to its full matching audio stem, retaining quiet conditioning outside the requested interval. \textbf{(b)} V2A reads the matching entity's visual features, then gates its update by the source timeline. The woman's V2A residual is zero while her A2V connection remains enabled. Arrows indicate information flow, and frame shading denotes enabled routing.}
    \label{fig:mask-routing}
\end{figure}

\noindent\textbf{Entity-routed Refinement.}
Stem Formation establishes source identity, timing, and a shared scene representation, but unrestricted cross-modal attention can still send one stem to the wrong visible source. Scene Integration re-noises the Stem Formation video, stem, and scene states and jointly denoises them under entity-routed cross-modal attention, restricting audio--visual information exchange according to each source's spatiotemporal mask and active interval (see Figure~\ref{fig:method-overview}).
These states provide the initial scene composition and source timing, while the added noise allows the backbone to improve video detail and audio naturalness. We guide this refinement by linking each visible entity to its own audio stem, with the detailed routing illustrated in Figure~\ref{fig:mask-routing}.

We segment each visible sound source in the Stem Formation video to obtain a spatiotemporal entity mask $M_i$ over its video tokens. A2V and V2A use this mask differently because their queries belong to different modalities. In A2V, a video query inside $M_i$ reads only the complete matching audio stem. Its quiet off-window tokens help guide the entity's inactive periods. In V2A, stem $i$ reads only video keys and values inside $M_i$. We multiply the resulting cross-modal residual by $m_i(\tau)$ so entity features guide the audio only during its requested interval. Appendix~\ref{app:entity-routing} specifies mask processing and routing for background and audio-only sources. We finally decode the refined video and source latents into one video and $N$ editable audio stems.

%% file: sections/experiments.tex
\section{Experiments and Applications}
\label{sec:experiments}

We evaluate how well \methodname generates distinct sounds that follow a requested timeline and belong to the same audiovisual scene. We first compare complete outputs, then through extensive ablations we isolate the contributions of carrier replay, source-specific guidance, scene broadcast, and entity routing. Finally, we demonstrate editing capabilities through the generated stems.

\paragraph{Methods and scenes.}
We compare \methodname with five baselines. Standard LTX-2.5 (~\cite{ltx_2}) generates a single audio track containing all sound sources. Naive batch expansion uses the same frozen LTX-2.5 to generate separate source-conditioned stems with a shared video, but without our proposed controls. We also compare against Ovi (~\cite{low2025ovi}), MiniMax H3~\cite{minimax2026h3}, and Gemini Omni (~\cite{google2026geminiomni}), which also generate all sound sources together in a single audio track.
We evaluate 15 scenes with 40 sources, including dialogue, music, animal sounds, and object interactions. We use the same three seeds per scene for LTX-2.5, naive batch expansion, Ovi, MiniMax H3, and \methodname, while Omni provides one output per scene. Results are averaged first within each scene and then across scenes. Appendix~\ref{app:benchmark}
provides settings, scene descriptions, and source schedules.

\paragraph{Evaluation.} We evaluate control based on event timing and source content. For speech content, requested-window word error rate (WER) compares each speech segment with the recognized words within its time window. Window success measures whether an event occurs within its allowed time interval, with up to 5\% of its activity allowed outside the interval, allowing a small tolerance for imprecise detected boundaries. Full-clip sound success measures whether a requested sound is detected when timing is unrestricted. Order accuracy measures whether events occur in the correct sequence, while off-window activity averages the fractions of speech and sound activity outside their allowed intervals. For source content and separation, source-assignment accuracy measures whether each stem better matches its intended source than other sources in the scene. Target CLAP measures similarity to the intended source description, while competing CLAP measures the highest similarity to another source description in the scene.

Single-track generation is evaluated with post-hoc audio separation as an alternative to generating separate stems. For speech, we use automatic speech recognition (ASR) and word-level temporal alignment on the mixed audio, which we found more reliable than separating speech with SAM Audio~\cite{shi2025samaudio}. For non-speech sounds, SAM Audio extracts each requested source for activity and timing evaluation. We evaluate speech after post-hoc separation and find no improvement. Overall, post-hoc separation cannot enforce the requested timeline, and interference from jointly generated sources remains in the outputs. See Appendices~\ref{app:evaluation-protocols} and~\ref{app:additional-results} for full protocols and results.

\vspace{-10pt}
\subsection{Generation Results}
\label{sec:generation-comparison}
\label{sec:qualitative-results}
\vspace{-5pt}
As an example, we use Hospital Farewell, a scene where two people have an emotional exchange beside an empty hospital bed. Appendix Figure~\ref{fig:hospital-results} compares our method with naive batch expansion in temporal control, source separation, speaking activity, emotional delivery, and acoustic consistency.

\vspace{-5pt}
\paragraph{Temporal control.}
The woman speaks at 1.0--3.8s and the man replies at 5.0--8.2s, with a pause between.
 \methodname places both turns within these windows, whereas naive batch expansion delays both voices until near the end of the clip, causing them to overlap. Across the benchmark, \methodname achieves \textbf{93.1\%} window success, compared with 15.0\% for naive expansion and 85.0\% for Omni (Table~\ref{tab:main-comparison}), indicating that most events remain within their allowed intervals. Relative to naive expansion, off-window activity falls from 69.7\% to \textbf{1.7\%}, showing that unwanted activity outside the requested windows is also substantially reduced. These results show that our timing controls preserve the requested audiovisual structure by coordinating source activity.

\vspace{-5pt}
\paragraph{Source content and separation.}
Each voice must remain in its own stem and match the correct speaker. In Hospital Farewell, naive batch expansion places the man's voice in both stems and fails to animate the woman. Our source-specific guidance separates competing voices, while entity routing assigns each voice to the intended speaker.
Across the benchmark, source-assignment accuracy rises from 82.2\% to \textbf{90.0\%} (Table~\ref{tab:main-comparison}), indicating less confusion between sources. For sounds with unrestricted timing, \methodname maintains separation between overlapping sources while achieving \textbf{100.0\%} full-clip sound success. 
Listeners prefer our separation in \textbf{\MainFullSep{}\%} of judgments after splitting ties equally (Table~\ref{tab:human-comparison}).
These results support audio separation, while the routing ablation below evaluates whether sounds activate the correct visible speaker.

\paragraph{Naturalness and scene fit.}
Correct words and speaking turns alone do not make the conversation feel natural. 
In Hospital Farewell, naive batch expansion produces flat, acoustically inconsistent voices that weaken the shared setting. Soundwich better captures the woman’s grief and man’s regret while maintaining consistent acoustic character.
Across the main study, \methodname ranks first for naturalness in \textbf{\MainFullNatFirst{}\%} of judgments, with a mean rank of \MainFullNat{} versus \MainLTXNat{} for standard LTX-2.5 and \MainNaiveNat{} for naive expansion (Table~\ref{tab:human-comparison}). These lower ranks show a preference for our outputs, highlighting the importance of shared context for coherent conversations.

\begin{table}[!tb]
    \centering
    \caption{\textbf{Joint generation with source-level control.}
    Comparison of temporal adherence, ordering, off-window activity, and source assignment for 15 scenes. Soundwich achieves the highest success for both windowed and full-clip sound requests and substantially reduces off-window activity. Source assignment is evaluated  for methods producing separate stems; N/A denotes mixed-only outputs.}
    \label{tab:main-comparison}
    \resizebox{\linewidth}{!}{%
        \begin{tabular}{@{}lrrrrrr@{}}
            \toprule
            Method & \makecell{Window\\WER $\downarrow$}
            & \makecell{Window\\success $\uparrow$}
            & \makecell{Full-clip sound\\success $\uparrow$}
            & \makecell{Order\\accuracy $\uparrow$}
            & \makecell{Off-window\\activity $\downarrow$}
            & \makecell{Source\\assignment $\uparrow$}\\

            \midrule
            Standard LTX-2.5 & 0.622 & 20.4\% & 83.3\% & \textbf{100.0\%} & 36.5\% & N/A\\
            Naive batch expansion & 0.779 & 15.0\% & 93.3\% & 79.3\% & 69.7\% & 82.2\%\\
            Ovi & 0.552 & 24.1\% & 76.7\% & 97.6\% & 39.2\% & N/A\\
            MiniMax H3 & 0.440 & 27.0\% & 90.0\% & 96.8\% & 32.2\% & N/A\\
            Gemini Omni & 0.072 & 85.0\% & \textbf{100.0\%} & 97.4\% & 10.7\% & N/A\\
            \methodname & \textbf{0.060} & \textbf{93.1\%} & \textbf{100.0\%}
            & \textbf{100.0\%} & \textbf{1.7\%} & \textbf{90.0\%}\\
            \bottomrule
        \end{tabular}%
    }
\end{table}

\begin{SCtable}[1][!tb]
    \caption{\textbf{Human evaluation.} Naturalness and timeline ranks range
    from 1 (best) to 3. Separation preference splits ties equally;
    N/A denotes a mixed-only output.}
    \label{tab:human-comparison}
    \centering
    \footnotesize
    \setlength{\tabcolsep}{2pt}
    \renewcommand{\arraystretch}{1.10}
    \begin{tabularx}{0.6\linewidth}{@{}Xccc@{}}
        \toprule
        Method & \makecell{Naturalness\\rank $\downarrow$}
        & \makecell{Timeline\\rank $\downarrow$}
        & \makecell{Separation\\preference $\uparrow$}\\
        \midrule
        Standard LTX-2.5 & \MainLTXNat{} & \MainLTXTime{} & N/A\\
        Naive batch expansion & \MainNaiveNat{} & \MainNaiveTime{} & \MainNaiveSep{}\%\\
        \methodname & \textbf{\MainFullNat{}} & \textbf{\MainFullTime{}} & \textbf{\MainFullSep{}\%}\\
        \bottomrule
    \end{tabularx}
\end{SCtable}

\vspace{-8pt}
\subsection{Ablation Studies}
\label{sec:ablation-results}
\vspace{-8pt}

\paragraph{Carrier replay controls event timing.}
To determine whether carrier replay is needed to enforce the requested schedule, we compare paired outputs with and without replay across 15 scenes. Without replay, source prompts specify what to generate but provide insufficient control over when sounds occur. Enabling replay raises window success from 15.6\% to 93.3\% and order accuracy from 73.3\% to 100.0\%, while reducing off-window activity from 63.06\% to 1.23\% (Table~\ref{tab:component-ablation} (Left)). Thus, more events occur within their allowed windows and in the correct order, with substantially less activity between them. The quiet carrier contributes to this suppression by providing features that guide the model toward silence. Replacing it with zero removes that guidance, allowing the model's learned prior to produce unwanted sound (Appendix~\ref{app:zero-vs-quiet}). Together, these comparisons support replaying both active and quiet features to control when each source sounds and when it remains silent.

\vspace{-5pt}
\paragraph{Source-specific guidance separates associated sounds.}
Each stem receives its own text condition to specify its sound and prevent other sources in the shared video from leaking through V2A attention. Our method does not rely on timing for separation. Across six scenes with fully overlapping sources and no timing constraints, source-specific guidance improves assignment accuracy from 61.1\% to 95.8\%, reduces competing-source CLAP from 0.262 to 0.115, and increases target-source CLAP from 0.313 to 0.344 (Table~\ref{tab:component-ablation} (Middle)). Thus, source-specific guidance improves separation even with fully overlapping sources, while timing provides additional control over source activity. See Appendix~\ref{app:negative-cfg-diagnostic} for examples and details.

\vspace{-5pt}
\paragraph{Scene broadcast improves contextual naturalness.}
While source-specific guidance keeps sounds separate, the scene stem provides the shared context needed for their emotional delivery and acoustic character to fit the same scene. We evaluate this role by comparing Scene Integration with and without scene broadcast. Adding broadcast raises the percentage of first-place rankings for contextual naturalness from \NoBroadcastContextFirst{}\% to \FullContextFirst{}\% and for overall preference from \NoBroadcastOverallFirst{}\% to \FullOverallFirst{}\% (Table~\ref{tab:component-ablation} (Right)). This preference indicates that separating sources is not enough: they also need shared context to form a coherent interaction. Stem Formation audio serves as a sound-quality reference, with refinement raising its score from \StageOneQuality{} to \NoBroadcastQuality{} without broadcast and \FullQuality{} with broadcast (Table~\ref{tab:broadcast-full-results}). The scene stem therefore helps reconcile distinct source content with coherent delivery.

\vspace{-5pt}
\paragraph{Entity routing preserves visual ownership.}
While scene broadcast helps sounds fit the same context, entity routing connects each sound to the movement of its intended visual source. Without routing, a sound can cause the wrong entity to move while its intended source remains inactive. Across the ten scenes in the human study, enabling routing raises correct-ownership responses from \NoRoutingOwnership{}\% to \FullOwnership{}\% and target-activation responses from \NoRoutingActivation{}\% to \FullActivation{}\%. These gains show that the intended source is more likely to become active and that its movement is more consistently associated with the correct sound. Entity routing therefore helps preserve the correspondence between separate audio stems and visible actions. We provide the evaluation questions and paired results in Appendix~\ref{app:routing-evaluation}.

\subsection{Applications}
\label{sec:editing-comparison}
\vspace{-5pt}

Separate stems let users change the timing or content of a selected sound while preserving audio they want to keep. The independent editing section in our supplementary website\footnote{\url{https://soundwich.avdemo.workers.dev/}} presents three before-and-after examples. In a scene where a cat meows during a conversation in an archive, we move the archivist's line earlier and the meow later while retaining the existing voices, words, and cat sound. In a dialogue between a woman and a detective, we replace the detective's reply while preserving the woman's original line and timing. In a scene where two people introduce themselves, we reverse who speaks first while keeping both recorded introductions intact. We describe the editing operations and refinement procedure in Appendix~\ref{app:stem-editing}.

We also show Soundwich for image-to-video generation using images from well-known animations in the supplementary video. Given an input image, Soundwich generates a video together with separate audio stems for its sound sources. This layered representation makes the generated result directly editable: we can swap dialogue between characters, change when they speak, remove individual dialogue, or remove background music while preserving the remaining audio and video.

\begin{table}[!tb]
    \centering
    \caption{\textbf{Contributions of timing carriers, source-specific guidance, and scene broadcast.}
    Left: automatic timing evaluation on 15 scenes. Middle: source separation from six scenes on concurrent sources with unrestricted timing. Target and competing CLAP measure similarity to the intended and strongest competing source descriptions. Right:
    Scene Integration broadcast comparison with \AblationTrials{} human listening judgments per condition from \AblationResponses{} completed responses. Naturalness and preference report the percentage of first-place rankings. Quality is scored from 0 to 10.}
    \label{tab:component-ablation}
    \fontsize{8}{10}\selectfont
    \begin{minipage}[t]{0.33\linewidth}
        \vspace{0pt}
        \fontsize{8}{10}\selectfont
        \centering
        \hrule
        \smallskip
        \textbf{\strut Timing carriers}\par\smallskip
        \setlength{\tabcolsep}{2pt}
        \begin{tabular*}{\linewidth}{@{\extracolsep{\fill}}lrr@{}}
            \toprule
            Measure & Off & On\\
            \midrule
            Window success $\uparrow$ & 15.6\% & \textbf{93.3\%}\\
            Order accuracy $\uparrow$ & 73.3\% & \textbf{100.0\%}\\
            Off-window activity $\downarrow$ & 63.06\% & \textbf{1.23\%}\\
            \bottomrule
        \end{tabular*}
    \end{minipage}\hfill
    \begin{minipage}[t]{0.33\linewidth}
        \vspace{0pt}
        \fontsize{8}{10}\selectfont
        \centering
        \hrule
        \smallskip
        \textbf{\strut Source-specific guidance}\par\smallskip
        \setlength{\tabcolsep}{2pt}
        \begin{tabular*}{\linewidth}{@{\extracolsep{\fill}}lrr@{}}
            \toprule
            Measure & Off & On\\
            \midrule
            Target CLAP $\uparrow$ & 0.313 & \textbf{0.344}\\
            Competing CLAP $\downarrow$ & 0.262 & \textbf{0.115}\\
            Source assignment $\uparrow$ & 61.1\% & \textbf{95.8\%}\\
            \bottomrule
        \end{tabular*}
    \end{minipage}\hfill
    \begin{minipage}[t]{0.30\linewidth}
        \vspace{0pt}
        \fontsize{8}{10}\selectfont
        \centering
        \hrule
        \smallskip
        \textbf{\strut Scene broadcast}\par\smallskip
        \setlength{\tabcolsep}{2pt}
        \begin{tabular*}{\linewidth}{@{\extracolsep{\fill}}lrr@{}}
            \toprule
            Measure & Off & On\\
            \midrule
            Naturalness $\uparrow$ & \NoBroadcastContextFirst{}\% & \textbf{\FullContextFirst{}\%}\\
            Quality score $\uparrow$ & \NoBroadcastQuality{} & \textbf{\FullQuality{}}\\
            Overall preference $\uparrow$ & \NoBroadcastOverallFirst{}\% & \textbf{\FullOverallFirst{}\%}\\
            \bottomrule
        \end{tabular*}
    \end{minipage}
\end{table}

%% file: sections/conclusion.tex
\vspace{-5pt}
\section{Conclusion, Limitations and Future Work}
\vspace{-5pt}

We have presented a training-free technique that turns the monolithic audio output into multiple synchronized and independently controllable sound layers.
Beyond the specific audio-video setting, our work suggests a broader idea: \emph{layers can serve as handles for controlling generative content}. Rather than treating the generated output as a single inseparable signal, exposing its constituent sources as separate layers creates explicit handles for controlling, editing, and recombining the generated content. We believe this idea may extend beyond sound to other generative settings or media in which several components must remain jointly coherent while retaining individual control.

A second interesting aspect of our approach is the use of the batch dimension to create these layers. Rather than modifying the architecture to explicitly increase the dimensionality of its output, we use the batch dimension to instantiate multiple parallel trajectories, coupled through a shared generative process. In this sense, the batch dimension becomes a mechanism for introducing structure into an otherwise monolithic output. This may offer a useful design principle for introducing layered representations into other pretrained generative models.

Our formulation remains challenging for complex scenes with many simultaneous or ambiguous sources, where interference between layers may increase. It also inherits the limitations and biases of the underlying pretrained model. A promising direction is to automatically discover sound layers and infer their temporal structure, reducing the need for explicit source and timing specifications. Extending Soundwich to longer videos is also important, but would require dynamic audio layers, as sources may naturally enter and leave the scene over time. Automatically discovering these sources and their temporal structure could enable layered generation beyond a fixed set of predefined stems.

%% file: sections/appendix_supplementary.tex
\section{Supplementary Material}
\label{app:supplementary-Material}

Additional audiovisual results and interactive comparisons are available
on our anonymous supplementary website:
\url{https://soundwich.avdemo.workers.dev/}.

%% file: sections/appendix_method_details.tex
\section{Detailed Method and Implementation}
\label{app:method-details}

We provide implementation details for the generation procedure
described in Section~\ref{sec:method}. We first summarize the complete generation procedure, then describe source conditioning, reference capture and timeline replay, scene gathering and broadcast, and entity routing, followed by model choices and inference settings.

\subsection{Generation Procedure}
\label{app:gen-procedure}
\methodname takes a visual prompt describing the scene and a separate specification for each sound source. Each source specifies the desired sound, competing sounds to exclude, and the intervals during which it should be audible. For sources associated with a visible entity, a text description identifies the entity to locate in the video. The output is one video and a separately editable audio track for each source, which we call a \emph{source stem}. Generation proceeds as follows.

\begin{enumerate}
    \item \textbf{Prepare reusable sound and silence references.} Before generating the target scene, we generate references containing the desired sound categories and separate references containing silence. We summarize selected intermediate audio features from these references into vectors called \emph{carriers}. These vectors provide reusable signals for encouraging sound or quiet at specified times.

    \item \textbf{Generate the initial video and source stems.} Stem Formation jointly generates one shared video and the separate source stems. Each stem receives its own sound description and exclusions to encourage the requested source and discourage competing sounds. During generation, the sound carriers encourage activity inside the requested intervals, while the silence carriers suppress activity outside them.

    \item \textbf{Collect shared scene information.} Alongside the source stems,
Stem Formation maintains an additional internal audio representation called the \emph{scene stem}. It gathers information from the source stems during their active intervals, collecting their timing and shared context. The source stems do not directly read from one another during this operation.

    \item \textbf{Locate the visible sound sources.} We apply a segmentation model to the video produced by Stem Formation using each visible source's text description. This produces masks identifying the region occupied by each entity over time. We inspect these masks before using them to associate the audio stems with their corresponding visible sources.

    \item \textbf{Refine the video and audio together.}
Scene Integration increases the spatial resolution of the video representation and adds noise to the saved video, source, and scene representations before jointly refining them. Each source stem reads context from the scene stem during its active intervals to support coherent interactions. The entity masks restrict information exchange between each visible source and its corresponding audio stem. Silence carriers continue to suppress sound outside the requested intervals.

    \item \textbf{Produce editable outputs.} We decode the refined representations into the final video and individual audio stems. The stems are retained separately and combined at their specified volumes to form the mixed soundtrack. The internal scene stem is discarded.
\end{enumerate}

\subsection{Inputs, Shared Video, and Guidance}
\label{app:source-guidance}

\paragraph{Source specification.}
Each source is described by a positive audio prompt, an optional exclusion prompt, activity intervals, a category-matched activation reference, a mixing gain controlling its volume, and, for visible sources, a segmentation prompt. Exclusions default to the other scene sources; an explicit prompt replaces that list, while global audio-quality exclusions always apply. Multiple disjoint activity intervals are supported; a source without specified intervals remains temporally unconstrained. Offscreen sounds and background music do not require a visible entity, and sources can also be selected individually to participate in gathering and receiving shared scene context.

\paragraph{Modality-specific guidance.}
Each audio stem should follow its own sound description and exclude competing sources, while the shared video should follow the complete scene description. We therefore use source-specific prompts for audio guidance and scene-level prompts for video guidance. To compute these two guidance updates, Stem Formation uses four forward passes through the joint audio–video model at each denoising step. Each pass processes the shared video and all audio stems together, producing predictions for both audio and video. We select the predictions needed for each guidance update as follows.

The first two passes compute audio guidance. Each source stem receives its positive audio prompt in the first pass and its negative audio prompt in the second. The video stream receives the positive visual prompt in both passes. We retain the audio predictions and discard the video predictions. The remaining two passes compute video guidance. The positive visual prompt conditions both audio and video in the third pass, and the negative visual prompt conditions both in the fourth. We retain the video predictions and discard the audio predictions.

Let $D_i^{a,+}$ and $D_i^{a,-}$ denote the retained positive and negative clean audio predictions, and $D^{v,+}$ and $D^{v,-}$ the corresponding video predictions. Classifier-free guidance combines them as
\begin{equation}
\begin{aligned}
\widehat D_i^a &= D_i^{a,+} + (s_a-1)\bigl(D_i^{a,+}-D_i^{a,-}\bigr),\\
\widehat D^v &= D^{v,+} + (s_v-1)\bigl(D^{v,+}-D^{v,-}\bigr).
\end{aligned}
\label{eq:appendix-cfg}
\end{equation}
We use $s_a=7$ and $s_v=3$. The pass count is independent of the number of sources, although computation per pass increases with source count. Scene Integration uses $s_a=s_v=1$ and one positive-conditioned pass per step.

\paragraph{Shared video computation.}
One video latent is shared by $N$ source-audio latents and one scene stem. Within each pass, video self-attention, text cross-attention, and feed-forward layers are evaluated once and reused across stems. A2V concatenates the source-audio keys and values along the token sequence while preserving each stem's temporal positions; the scene stem is excluded. In V2A, each audio state attends to the shared video. Source audio self-attention remains independent, with scene context exchanged through gathering and broadcast (Appendix~\ref{app:scene-sharing}). Scene Integration further restricts A2V and V2A using entity masks (Appendix~\ref{app:entity-routing}).

\begin{table}[tbp]
    \centering
    \caption{Sounding- and quiet-reference prompts and random seeds. The positive and negative columns give the stored reference conditions.}
    \label{tab:carrier-prompts}
    \small
    \setlength{\tabcolsep}{4pt}
    \begin{tabular}{@{}p{0.13\textwidth}p{0.44\textwidth}p{0.30\textwidth}r@{}}
    \toprule
    Source & Positive condition & Negative condition & Seed \\
    \midrule
    Female speech & high quality clear female voice speaking loudly & male voice & 42 \\
    Male speech & male voice speaking loudly & female voice & 42 \\
    Cat meow & A cat meows loudly and repeatedly. & speech, music, silence, quiet sound & 33 \\
    Violin & A violin plays loudly and continuously. & speech, noise, silence & 33 \\
    Piano & A piano plays loudly and continuously. & speech, noise, silence & 33 \\
    Applause & people clapping loudly and repeatedly & laughter, music & 33 \\
    Door knock & A person knocks on a wooden door with loud knocks. & footsteps, speech, music & 41 \\
    Footsteps & A person walks on a wooden floor with loud footsteps. & door knocking, speech, music & 41 \\
    Guitar & An acoustic guitar plays loudly and continuously. & speech, noise, silence & 33 \\
    Dog bark & A dog barks loudly and repeatedly. & speech, music, silence, quiet sound & 31 \\
    Quiet & A uniform pure black frame remains perfectly still. Audio: digital silence, inaudible noise floor & camera cuts, text, blur, low quality. Audio: speech, music, singing, loud sound, random noise & 101 \\
    \bottomrule
    \end{tabular}
\end{table}

\subsection{Reference Capture and Timeline Replay}
\label{app:cached-carriers}
To place each sound within its requested intervals, we reuse audio features captured from references containing either the desired sound or silence. These features provide activation carriers that encourage sound inside the intervals and suppression carriers that encourage quiet outside them. We describe the reference generation, carrier capture, and replay procedure below.

\paragraph{Reference generation.}
We generate category-specific sounding references, including separate male and female speech references, and one shared quiet reference using the prompts and seeds in Table~\ref{tab:carrier-prompts}. Stem Formation references use the same model, denoising schedule, and CFG scales as target generation (Appendix~\ref{app:generation-settings}). References are generated in advance and reused across scenes, utterances, and evaluation seeds; assignments to sources were manually selected before the main comparison.

\paragraph{Carrier capture.}
At every denoising step and transformer block, we record audio self-attention outputs from the positive-conditioned reference pass. For each token vector $h_t\in\mathbb{R}^d$, its magnitude is the channel RMS, $e_t=(d^{-1}\sum_{j=1}^{d}h_{t,j}^2)^{1/2}$. Averaging the top 30\% of sounding-reference vectors gives the \emph{activation carrier}; averaging the bottom 30\% of quiet-reference vectors gives the \emph{suppression carrier}. Each carrier is a single vector without reference-time ordering, stored at unit RMS in FP16 together with its original RMS. Sources use category-matched activation carriers and share the quiet carriers.

\paragraph{Timeline replay.}
To control when each source is audible, we apply activation carriers during its requested intervals and suppress sound outside them by weakening the existing audio features and blending them with quiet-reference features. For source $i$, an activity gate $m_i$ controls the transition between these operations. It equals one inside each requested interval, where activation is applied and suppression is disabled. To avoid abrupt changes, we extend each boundary with a $0.20$-second linear transition. Beyond these transitions, activation is disabled ($m_i=0$), and suppression is fully applied. For multiple intervals, we construct a gate for each interval and take their maximum at each audio position. Sources without specified activity intervals bypass carrier replay.

During Stem Formation, these operations modify the audio self-attention output at every transformer block in the positive audio pass. Let $h_i$ denote the current self-attention output for source $i$, and let $s_{n,\ell}$ denote the quiet carrier captured at denoising step $n$ and transformer block $\ell$, restored to its original RMS magnitude. We define the suppression weight as $o_i=1-m_i$ and apply the two operations in sequence:
\begin{equation}
h'_i=(1-0.25o_i)h_i,\qquad
h''_i=(1-0.60o_i)h'_i+0.60o_i s_{n,\ell}.
\end{equation}
Where suppression is fully active ($m_i=0$), the resulting features are $0.30h_i+0.60s_{n,\ell}$. Inside the requested intervals ($m_i=1$), both operations leave $h_i$ unchanged. We then add the scaled activation carrier to encourage the requested sound at the specified times. Equation~\ref{eq:carrier-replay} summarizes the combined suppression and activation update.

\paragraph{Replay scaling and schedules.}
After placing the activation carrier over the requested intervals, we measure its RMS and the source-feature RMS within those intervals. We scale the carrier by their source-to-carrier RMS ratio, capped at four times amplification, then apply the activation strength and activity gate before addition. The effective activation strength is $0.40$ for the first 10\% of Stem Formation denoising and $0.08$ thereafter; quiet carriers retain their recorded low RMS.

Scene Integration applies off-window suppression without activation replay, using quiet carriers captured with the distilled model at the matching refinement step and block. Table~\ref{tab:method-settings} collects the settings. Appendices~\ref{app:zero-vs-quiet} and~\ref{app:activation-direction} compare quiet versus zero suppression targets and activation profiles within an interval, respectively.

\subsection{Scene Gather and Broadcast}
\label{app:scene-sharing}
Separately generated source stems can follow their individual prompts while lacking the shared context of a conversation or performance. We use the internal scene stem to collect information from the active sources during Stem Formation and make that information available to each source during Scene Integration. Gathering and broadcast provide this exchange while keeping the source stems’ audio self-attention separate.

\begin{figure}[tbp]
    \centering
    \includegraphics[width=\linewidth]{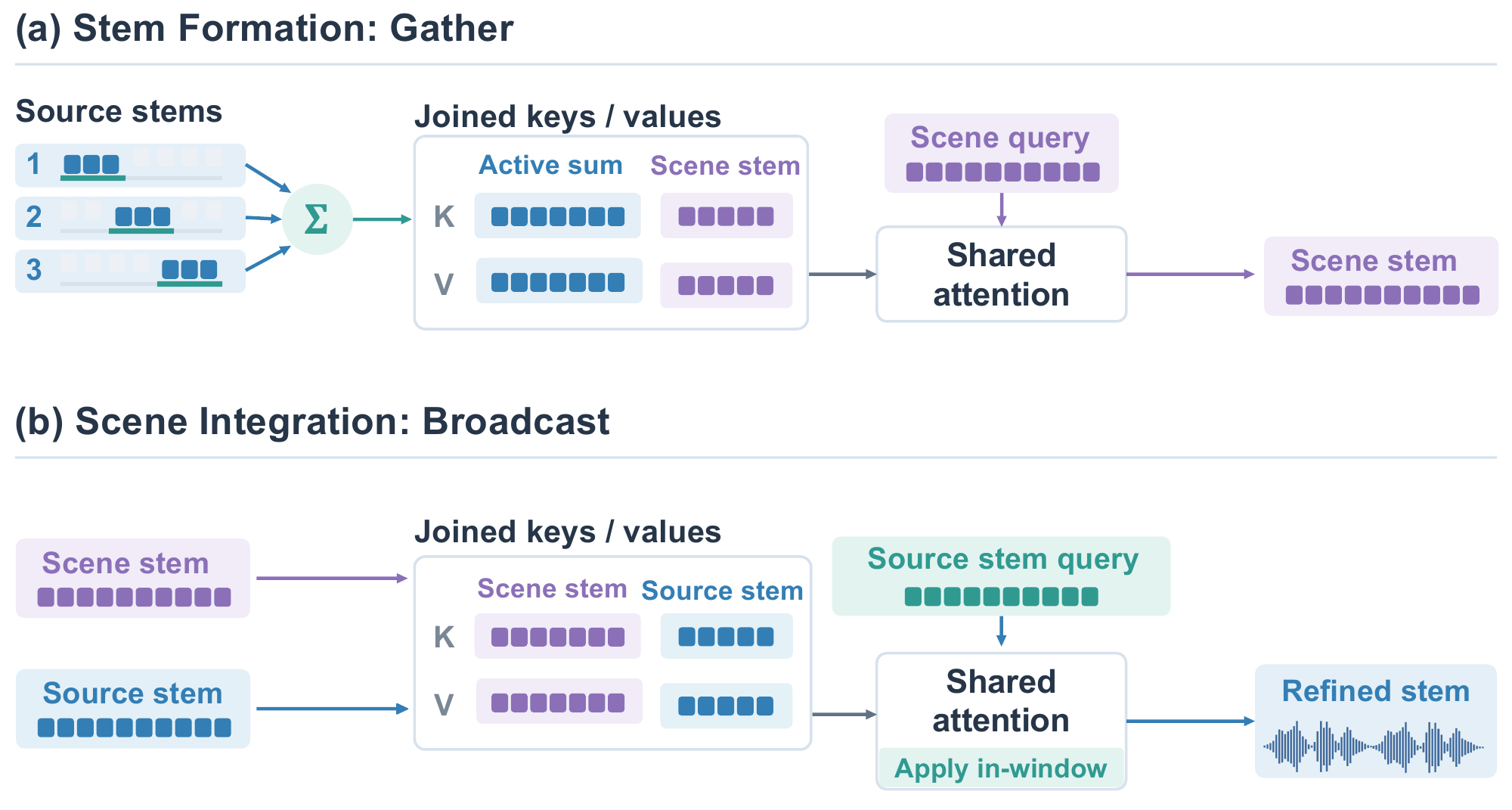}
    \caption{\textbf{Gather--broadcast attention.} \textbf{(a)} Active portions of source states are summed before projection. The scene stem queries its own keys/values and those of the aggregate, while source self-attention stays independent. \textbf{(b)} Each participating source stem queries its own and the evolving scene stem's keys/values. The shared-attention update is applied within the source's active interval. The scene latent carries context from Stem Formation to Scene Integration. }
    \label{fig:gather-broadcast}
\end{figure}

\paragraph{Gathering.}
For participating sources $\mathcal E$, let $u_i(\tau)$ be the current audio features at the sharing step and block, and $m_i(\tau)$ the activity gate. Before key/value projection, we form
\begin{equation}
r(\tau)=\sum_{i\in\mathcal E}m_i(\tau)u_i(\tau).
\end{equation}
The scene queries attend to their own keys and values and those of $r$, while source self-attention remains independent (Figure~\ref{fig:gather-broadcast}(a)).

Gathering replaces the scene's self-attention output in the positive audio pass during the first half of Stem Formation denoising; ordinary self-attention resumes in the second half. Quiet-carrier suppression applies outside the union of participating sources' activity intervals.

The raw sum preserves source magnitudes. The RMS-balanced variant normalizes each source to unit RMS before gating and summing, then multiplies by the mean source RMS. Appendix~\ref{app:benchmark} identifies the variant used for each benchmark configuration. Participation is selected per source, allowing persistent background music to bypass scene sharing while remaining an output stem. The scene prompt combines the visual description, participating-source descriptions, and global audio-quality condition.

\paragraph{Refinement initialization.}
Scene Integration starts from the saved source-audio and scene states and the video state spatially upscaled to the target resolution. Each state $z_0$ is mixed with Gaussian noise $\epsilon\sim\mathcal N(0,I)$ at $\sigma_*=0.95$:
\begin{equation}
z_{\sigma_*}=(1-\sigma_*)z_0+\sigma_*\epsilon.
\label{eq:appendix-renoise}
\end{equation}
The shared video receives one noise sample, while each source and the scene receive independent samples. All states are refined jointly.

\paragraph{Broadcasting.}
During Scene Integration, each source uses the collected scene context to refine its sound as part of the same conversation or performance. We provide this context through audio self-attention: each participating source supplies its own queries and attends to both its own keys and values and those of the scene stem, as illustrated in Figure~\ref{fig:gather-broadcast}(b). This lets each source select information from the shared context while retaining its separate audio representation.

To keep this update on the source's existing feature scale, we first match the RMS magnitude of the attention output that includes scene context to that of the source's ordinary self-attention output. The attention output incorporating scene context replaces the ordinary output inside the requested intervals, with gradual transitions at their boundaries. We rescale the blended result to preserve the original source RMS. Sources excluded from the gathering also bypass the broadcast.

To keep the scene stem from drifting away from the sources during refinement, we recompute the timeline-weighted source mix defined in Gathering. We use this update to preserve the overall activity pattern of the sources while allowing the audio texture to improve during refinement. We match its RMS magnitude to that of the scene features, add it with strength $0.10$, and rescale the result to preserve the scene RMS. 

\subsection{Entity Routing and Audio Output}
\label{app:entity-routing}
To associate each audio stem with its visible source, we restrict audio--video attention using masks that locate the source in the video. These masks determine which audio stem can influence each video region and which visual features can contribute to each audio stem.

\paragraph{Mask processing and source tracking.}
For each full-method sample, SAM3.1~\citep{carion2026sam3} tracks visible sources in 32 frames sampled from the Stem Formation video at stride eight. Each source's segmentation prompt is applied at frame zero with score threshold $0.50$. We inspect tracks before Scene Integration and can correct regions with positive or negative point prompts. Accepted masks are resampled to the spatial and temporal video-token resolution; tokens with mask values exceeding $0.05$ are included.

\paragraph{Audio--video routing.}
Each visible source should respond to its own audio stem, including the periods when that stem is quiet. We therefore use A2V attention to connect video queries inside an entity's mask to its complete audio stem. Keeping this connection during quiet intervals lets the quiet audio features guide the entity's behavior when it should not produce sound. Figure~\ref{fig:mask-routing} illustrates this routing when the woman is quiet and the man is speaking.

Each audio stem should receive visual information from its own entity during the periods when that source should produce sound. We therefore use V2A attention to pass visual information from each entity's masked region to its audio stem during the requested activity intervals. Gating this update allows the entity's visual features to guide its sound only during the requested intervals and disables their contribution during quiet periods, with gradual transitions at the interval boundaries.

When entity masks overlap, A2V assigns each video token to the source with the largest mask value. Video queries outside all entity masks can attend to all source stems with valid masks. Audio-only sources and the internal scene stem are excluded from A2V. A source without a valid mask receives no routed V2A update, while the internal scene stem uses standard V2A without mask or timeline restrictions.

\paragraph{Decoding and mixing.}
We decode each refined source separately at 24~kHz with the audio VAE and apply LTX-2.5 bandwidth extension to 48~kHz, retaining individual stems. Before mixing, we scale each stem toward peak amplitude $0.95$, cap amplification at four times its original level, and leave stems with peaks below $10^{-3}$ unamplified. We then apply source-specific volumes and sum the stems. If the mixture's peak exceeds $0.95$, we scale it down to peak amplitude $0.95$.

\subsection{Implementation Settings and Model Assets}
\label{app:generation-settings}

\paragraph{Model selection.}
To fit generation within the 32~GB memory of our NVIDIA RTX~5090, we use 8-bit integer (INT8) quantized weights for the LTX-2.5 22B development and distilled models and the Gemma-4 12B text encoder. 

Our choice of model for Stem Formation and Scene Integration follows our experimental observations. High classifier-free guidance (CFG) scales were effective with the non-distilled development model but degraded generation quality with the distilled model. We therefore use the development model for
Stem Formation generation with source-specific guidance, and the distilled model for
Scene Integration refinement with CFG set to $1$.

\paragraph{Generation settings.}
Each generated video contains 249 frames at 25 fps, corresponding to 9.96 seconds.
Stem Formation operates at $384\times224$ resolution using the standard LTX-2.5 schedule with 40 Euler steps. Video and audio CFG scales are $3$ and $7$, respectively.

A $2\times$ latent spatial upscaler increases the video resolution to $768\times448$ before
Scene Integration. Refinement uses eight Euler steps with the noise-level sequence
\[
[0.95,\;0.8375,\;0.725,\;0.6240,\;0.5229,\;0.421875,\;0.28125,\;0.140625,\;0].
\]
The saved video, source-audio, and scene states are re-noised at the starting level $0.95$ and refined together, following Equation~\ref{eq:appendix-renoise}.
Scene Integration uses the positive audio and video predictions without additional CFG. The video VAE, audio VAE, and spatial upscaler use BF16 precision. The final audio stems are decoded from the refined source states and extended to 48~kHz.

\paragraph{Method settings.}
Table~\ref{tab:method-settings} collects the settings for carrier capture and replay, scene gathering and broadcast, and entity routing. Activation strengths include the fixed carrier value scale, giving effective strengths of $0.40$ for the first 10\% of
Stem Formation denoising and $0.08$ thereafter. The benchmark includes both raw and RMS-balanced source aggregation, with the configurations identified in Appendix~\ref{app:benchmark}.

\begin{table*}[t]
    \centering
    \caption{Inference settings for carrier control, scene coordination, and entity routing. Percentages of denoising refer to progress through
Stem Formation.}
    \label{tab:method-settings}
    \small
    \setlength{\tabcolsep}{5pt}
    \begin{tabular}{@{}p{0.19\textwidth}p{0.34\textwidth}p{0.39\textwidth}@{}}
        \toprule
        Component & Setting & Value \\
        \midrule
        Carrier capture & Activation / suppression selection & Top 30\% / bottom 30\% by RMS magnitude \\
        Carrier capture & Stored representation & One FP16 unit-RMS vector and its original RMS per step and block \\
        Activation replay & Effective strength & $0.40$ for the first 10\%, then $0.08$ \\
        Activation replay & RMS-matching multiplier cap & $4.0$ \\
        Suppression & Feature attenuation / quiet-carrier blend & $0.25$ / $0.60$ \\
        Timeline control & Boundary transition width & $0.20$ seconds \\
        Stem Formation\newline gathering & Schedule / attention strength & First 50\% of denoising / $1.0$ \\
        Stem Formation\newline gathering & Source aggregation & Raw or RMS-balanced sum \\
        Stem Formation\newline gathering & Direct source--scene feature additions & Disabled \\
        Scene Integration\newline broadcast & Attention strength & $1.0$ \\
        Scene Integration\newline scene update & Source-mix update strength & $0.10$ \\
        A2V routing & Spatial restriction / output gain & Entity mask / $1.0$ \\
        V2A routing & Spatial and temporal restrictions & Entity mask and source activity gate \\
        Entity masks & Video-token mask threshold & $0.05$ \\
        \bottomrule
    \end{tabular}
\end{table*}

\paragraph{Model assets.}
The development and distilled transformer files are \texttt{ltx-\allowbreak{}2.5-\allowbreak{}22b-\allowbreak{}dev-\allowbreak{}transformer-\allowbreak{}comfy-\allowbreak{}int8-\allowbreak{}convrot} and \texttt{ltx-\allowbreak{}2.5-\allowbreak{}22b-\allowbreak{}distilled-\allowbreak{}transformer-\allowbreak{}comfy-\allowbreak{}int8-\allowbreak{}convrot}. The text encoder is \texttt{gemma4-\allowbreak{}12b-\allowbreak{}with-\allowbreak{}proj-\allowbreak{}ltx-\allowbreak{}2.5-\allowbreak{}comfy-\allowbreak{}int8-\allowbreak{}convrot}. All three files use the \texttt{.safetensors} extension and contain prequantized weights. The video VAE, audio VAE, and spatial upscaler are the corresponding BF16 LTX-2.5 assets.

%% file: sections/appendix_benchmark.tex
\section{Experimental Setup}
\label{app:benchmark}

Implementation settings and evaluation protocols appear in Appendices~\ref{app:generation-settings} and~\ref{app:evaluation-protocols}.

\subsection{Scenes and Comparison Design}

The benchmark has 15 scenes, each with two to four sources: 40 total, comprising 27 speech and 13 musical or other sound sources. It covers ordered dialogue, timed events, continuous instruments, and overlapping sounds. Scene specifications include a visual description, desired and excluded source sounds, activity intervals, and visible-source segmentation prompts.

Prompts, carrier assignments, source volumes, and scene-context participation were fixed before the three-seed comparison. Continuous instruments bypass gathering and broadcast to avoid dominating the shared context; Neon biology laboratory's timed dog and cat sounds also bypass sharing to keep its scene context focused on dialogue. All remain editable output stems (Table~\ref{tab:benchmark-scenes}).

Three seeds per scene were sampled once with master seed 2027 and reused by standard LTX-2.5, naive batch expansion, Ovi, MiniMax H3, and \methodname. Table~\ref{tab:benchmark-seeds} combines the schedules and paired seeds.

\begin{table}[htbp]
    \centering
    \caption{Benchmark sources, schedules (seconds), and paired seeds. F/M denote female/male voices; full-length sources are requested throughout the clip. $\dagger$: excluded from gathering and broadcast, retained as editable output stems.}
    \label{tab:benchmark-scenes}
    \label{tab:benchmark-seeds}
    \scriptsize
    \setlength{\tabcolsep}{3pt}
    \begin{tabular}{@{}p{0.20\textwidth}p{0.43\textwidth}p{0.12\textwidth}p{0.20\textwidth}@{}}
        \toprule
        Scene & Scheduled sources & Full-length & Seeds 1, 2, 3 \\
        \midrule
        Noir office & F (0.6--4.0); M (4.4--8.0) & -- & 1775, 7604, 1304 \\
        Enchanted archive & cat meow (0.4--2.3); F (2.6--5.5); M (5.9--9.0) & -- & 223, 5284, 6654 \\
        Crystal court & F (1.0--4.0); M (5.0--8.0) & violin$^\dagger$ & 3007, 6534, 2951 \\
        Orbital greenhouse & F (0.8--4.1); M (4.5--8.0) & -- & 5698, 9754, 2934 \\
        Android laboratory & F (1.0--4.0); M (4.5--7.5) & -- & 4534, 4733, 9989 \\
        Astral conservatory & F (1.0--4.1); M (5.0--8.2) & violin$^\dagger$; piano$^\dagger$ & 1256, 5598, 4951 \\
        Neon biology laboratory & dog bark$^\dagger$ (0.4--2.8); cat meow$^\dagger$ (0.4--2.8); F (3.1--6.0); M (6.3--9.2) & -- & 1096, 4174, 4179 \\
        Orbital award ceremony & applause (0.3--2.8); F (3.0--5.9); M (6.2--9.3) & piano$^\dagger$ & 6968, 8290, 3744 \\
        Clockwork airship & door knock (0.4--2.8); footsteps (3.0--5.5); F (6.0--9.2) & guitar$^\dagger$ & 2565, 5532, 5671 \\
        Animated kitchen & F (0.5--3.2); M (3.6--6.4) & -- & 3596, 6522, 7436 \\
        Podcast studio & F (0.4--3.0); M (3.4--6.6) & -- & 6250, 2688, 7671 \\
        Caf\'e performance & F (2.5--5.8) & guitar$^\dagger$ & 6126, 5164, 2038 \\
        Animated presentation & F (0.5--3.6); M (4.0--7.0) & -- & 3137, 1014, 5329 \\
        Studio introduction & F (0.5--3.6); M (4.0--7.4) & -- & 7797, 4230, 8605 \\
        Living-room cat & cat meow (1.0--3.8); M (4.6--7.4) & -- & 1610, 4481, 6791 \\
        \bottomrule
    \end{tabular}
\end{table}

\subsection{Compared Methods and Output Collection}

All six methods use the same 15 scenes. Standard LTX-2.5, naive batch expansion, Ovi, MiniMax H3, and \methodname each provide three outputs per scene (45 per method); Gemini Omni provides one (15 total) since it's a commercial model, giving 240 outputs. Scores average evaluated outputs within each scene, then average scenes equally; Appendix~\ref{app:joint-audio-metrics} defines the speech/sound measurements and eligibility criteria.

\paragraph{LTX-2.5 methods.}
Standard LTX-2.5 uses a joint scene prompt to generate one mixed soundtrack. Naive batch expansion uses separate source audio prompts and states with shared video. \methodname adds source-specific guidance, carrier replay, scene gathering/broadcast, and entity routing. All use the same final video resolution. Standard LTX-2.5 uses its official pipeline's Stage 1 audio; naive expansion uses refined source stems, and \methodname uses Scene Integration stems, both mixed for comparison.

\paragraph{Ovi.}
The Ovi pipeline~\cite{low2025ovi} with pretrained Ovi~1.1 produces ten-second videos with mixed audio: 50 UniPC denoising steps, video and audio CFG scales of $4$ and $3$, skip-layer guidance (SLG) at layer 11, INT8 quantization, and CPU offloading. Ovi adjusts the requested $704\times1280$ (height$\times$width, pixels) resolution to its spatial grid. Prompts specify speech and sound-event intervals.

\paragraph{MiniMax H3.}
We use the pretrained \texttt{MiniMaxAI/MiniMax-H3} model~\cite{minimax2026h3} for text-to-audio-video generation with 32 denoising steps. Each output contains 243 frames at 24 frames per second, giving 10.125-second clips at $448\times768$ resolution (height$\times$width, pixels), with a mixed soundtrack. We additionally include the global soundscape description in the input. Evaluation uses the first ten seconds of each output.

\paragraph{Omni.}
We collected one video with mixed audio per scene from Gemini Omni~\cite{google2026geminiomni} on September 8--11, 2026, using benchmark XML specifications without method-specific control tags or selected random seeds. Records identify the service and access dates, but no backend revision.

\subsection{Experiment-Specific Generation Conditions}
\label{app:carrier-metrics}

\paragraph{Scene aggregation.}
The method description uses raw summation (Appendix~\ref{app:scene-sharing}); 39 of the 45 main-comparison \methodname outputs use RMS-balanced aggregation. The remaining six use raw sums: all three seeds for Animated kitchen and Neon biology laboratory.

\paragraph{Carrier comparison.}
Replay on/off variants use one paired seed per benchmark scene, covering 27 speech and 13 non-speech stems. We use the joint speech/sound measurements (Appendix~\ref{app:joint-audio-metrics}); off-window activity averages speech and sound off-window fractions over available sources within each output, then over scenes. All 40 sources are evaluable with carriers; without carriers, one Android laboratory speech source fails the word-alignment criterion, leaving 39. Window success and off-window activity include all 15 scenes in both conditions; order accuracy includes 14 with carriers and 12 without.

\paragraph{Source-specific guidance.}
To test whether source-specific guidance reduces competing sounds within stems, we evaluate Stem Formation audio from six selected concurrent-source scenes (Table~\ref{tab:additional-scenes}; comparison in Appendix~\ref{app:negative-cfg-diagnostic}).

\subsection{Additional Evaluation Scenes}
\label{app:component-scenes}

Table~\ref{tab:additional-scenes} groups additional scenes by experiment; component studies also reuse benchmark scenes as specified in their protocols.

\begingroup
\small
\setlength{\tabcolsep}{4pt}
\renewcommand{\arraystretch}{1.12}
\begin{longtable}{@{}>{\raggedright\arraybackslash}p{0.20\textwidth}>{\raggedright\arraybackslash}p{0.45\textwidth}>{\raggedright\arraybackslash}p{0.29\textwidth}@{}}
\caption{Additional evaluation scenes. W and M denote the woman and man. Times are requested activity intervals in seconds. ``Unscheduled'' means no explicit activity windows. Source-specific-guidance sources are concurrent and unscheduled.}\label{tab:additional-scenes}\\
\toprule
Scene & Setting and source content & Source schedule\\
\midrule
\endfirsthead
\caption[]{Additional evaluation scenes (continued).}\\
\toprule
Scene & Setting and source content & Source schedule\\
\midrule
\endhead
\midrule
\multicolumn{3}{r}{\emph{Continued on the next page}}\\
\endfoot
\bottomrule
\endlastfoot
\multicolumn{3}{@{}l}{\textbf{Contextual naturalness}}\\*
\midrule
        Rainy caf\'e & Woman hesitantly asks if her companion is leaving; he answers with quiet affection. & W: 1.0--4.0; M: 5.0--8.3\\
        Interrogation room & Detective presses witness on a contradiction; reluctant confession follows. & W: 0.4--2.6, 5.3--7.5; M: 2.9--5.0, 7.8--9.6\\
        Stalled elevator & Frightened female passenger struggles to breathe; her coworker offers steady reassurance. & W: 0.3--2.2, 5.0--7.2; M: 2.5--4.7, 7.5--9.8\\
        Escape room & Two players race to open a locked box: urgency/panic become celebration. & W: 0.2--2.4, 5.0--7.2; M: 2.7--4.7, 7.5--9.7\\
        Ferry ticket mix-up & Two travelers discover misbooked tickets: frustration becomes teasing and dry humor. & W: 0.15--2.25, 4.85--7.15; M: 2.45--4.6, 7.35--9.7\\
        Kiln reveal & Successful glaze: woman feels trembling delight; companion feels warm pride. & W: 0.9--3.7; M: 5.0--8.6\\
        Baking comeback & Collapsed cake: embarrassment, playful suggestion, amused relief. & W: 0.2--2.8, 6.8--9.1; M: 3.6--6.0\\
        Record reunion & Man recognizes old album with astonishment; woman replies with shy joy. & M: 0.2--3.4; W: 4.3--8.0\\
        Laundromat joke & Shrunken sweater: man's comic outrage, woman's sheepish amusement. & M: 0.6--3.5; W: 4.4--7.8\\
        Watch-shop argument & Customer and shop owner argue angrily over sale of a family watch. & M: 0.4--3.0, 6.55--9.45; W: 3.35--6.05\\
        Gallery reflection & Woman asks if painting feels sad; man offers calm reassurance. & W: 1.0--3.8; M: 5.0--8.2\\
        Art auction surprise & Unexpected sale price: stunned disbelief, amused confidence, delighted laughter. & W: 0.65--3.25, 7.15--9.65; M: 3.95--6.45\\
        Night-train homecoming & Returning traveler meets guarded astonishment, then a warm welcome. & W: 0.5--3.35, 7.2--9.75; M: 4.05--6.55\\
        Family-home sale & Woman confronts man over selling their childhood home; he explains and admits fear. & W: 0.5--2.7, 5.3--7.5; M: 3.0--5.0, 7.8--9.8\\
\midrule
\multicolumn{3}{@{}l}{\textbf{Qualitative example and study tutorials}}\\*
\midrule
Hospital farewell & Beside an empty hospital bed, woman asks ``Did you bring the watch'' with restrained grief; man replies ``Yes it was beside the bed'' with soft regret. & W: 1.0--3.8; M: 5.0--8.2. Naturalness tutorial and qualitative example.\\
Studio conversation & Close two-shot: male host says ``The concert begins at seven''; female guest replies ``I will meet you there.'' & Host: 0.5--3.4; guest: 3.8--6.8. Routing tutorial.\\
\midrule
\multicolumn{3}{@{}l}{\textbf{Entity routing}}\\*
\midrule
Market conversation & Busy market: three visible speakers exchange ``These came in this morning,'' ``I will take a dozen,'' and ``They smell wonderful''. & Seller: 0.5--3.2; buyer: 3.6--6.2; older woman: 6.6--9.2.\\
\midrule
\multicolumn{3}{@{}l}{\textbf{Source-specific guidance}}\\*
\midrule
Auto repair shop & Revving car engine, impact wrench, hammering on a body panel, radio music. & Four sources.\\
Construction site & Jackhammer, circular saw, reversing-truck beeps, radio music. & Four sources.\\
Parade square & Brass band, marching drums, crowd cheers, motorcycle engine. & Four sources.\\
Quiz show & Speaking host, house band, audience applause. & Three sources.\\
Theater curtain & Speaking actor, orchestral finale, audience applause. & Three sources.\\
Blues quartet & Electric guitar, tenor saxophone, upright piano. & Three output stems.\\
\midrule
\multicolumn{3}{@{}l}{\textbf{Direct cross-stem sharing}}\\*
\midrule
Speakers & Woman: ``The garden smells of roses''; man: ``My train leaves at seven.'' & Two unscheduled speech stems.\\
Musicians & Violin sustains a bowed phrase while acoustic guitar plays separated plucked notes. & Two unscheduled instrument stems.\\
Percussion & Brass handbell: long ringing decays; wooden block: short, dry knocks. & Two unscheduled percussion stems.\\
\midrule
\multicolumn{3}{@{}l}{\textbf{Activation direction}}\\*
\midrule
Quiet room & Man: ``The room is quiet''; static medium close-up. & Male voice: 0.5--9.5.\\
Reception desk & Woman holding parcel: ``Your parcel is right here''; static medium close-up. & Female voice: 0.5--9.5.\\
Quiet garden & Older man: ``The sky looks clear''; static medium close-up. & Male voice: 0.5--9.5.\\
\midrule
\multicolumn{3}{@{}l}{\textbf{Ovi transfer}}\\*
\midrule
Classroom & Teacher greets class; children answer together; teacher asks them to open books. & Teacher: 1.0--2.4, 6.0--7.5; children: 3.2--4.8.\\
Ticket counter & Passenger requests tickets and thanks clerk; clerk gives departure information. & Passenger: 1.0--2.8, 7.0--8.5; clerk: 4.0--5.8.\\
Auction & Auctioneer announces start time; woman confirms bidding card; man confirms payment account. & Auctioneer: 0.5--3.2; W: 3.5--6.2; M: 6.5--9.3.\\
\end{longtable}
\endgroup

%% file: sections/appendix_evaluation_protocols.tex
\section{Evaluation Protocols and Validation}
\label{app:evaluation-protocols}

Benchmark scenes and generation conditions are in Appendix~\ref{app:benchmark}.

\subsection{Joint Speech and Sound Comparison}
\label{app:joint-audio-metrics}

We evaluate requested speech content, activity windows, and event order over the first ten seconds of each output.

\paragraph{Speech content and timing.}
Whisper \texttt{large-v3-turbo} transcribes individual speech stems when available, otherwise the mixed soundtrack; \texttt{wav2vec2-base-960h} estimates recognized words' start and end times. Requested-window word error rate (WER) compares words whose midpoints lie inside each requested interval with its requested text, counting substitutions, omissions, and insertions relative to the requested word count. We average WER over intervals within tracks, then tracks within outputs.

For timing, we locate each requested utterance in the transcript using word-sequence similarity of at least 0.55 (1 is exact), allowing transcription errors. Matched words' start and end times define utterance activity.

\paragraph{Non-speech activity.}
We use generated source stems for multi-stem methods; for Standard LTX-2.5, Ovi, MiniMax H3, and Omni, SAM Audio~\cite{shi2025samaudio} extracts each source from the mixture using its description. Activity detection thresholds root-mean-square (RMS) amplitude in 20-ms windows at 10-ms hops: the threshold is 10\% of the track's 95th-percentile RMS, floored at 0.001. We group consecutive active windows, bridge gaps of at most 100~ms, and discard detections shorter than 60~ms. Detected intervals are paired with requested events in temporal order, maximizing matches and breaking ties by closest start times.

\paragraph{Off-window activity.}
Off-window speech is recognized-word duration outside allowed intervals divided by total recognized-word duration, including extra and repeated speech. Allowed intervals belong to the source for individual stems and comprise all speakers' requested intervals for mixtures. Off-window sound analogously divides detected active duration outside the source's intervals by total active duration. The combined score averages speech and sound fractions over available sources within outputs, outputs within scenes, then scenes.

\paragraph{Window success and event order.}
Events may start anywhere within, and occupy part or all of, their requested interval. Success requires at least 95\% of detected activity inside it: summed matched-word durations for speech, or all detected active portions of a non-speech track. Missing detections count as failures. Window success is the fraction of scorable events satisfying this criterion.

Order accuracy uses matched-event pairs with non-overlapping requested intervals, including speech--sound pairs; overlapping requests permit either onset order. A pair is correct only if the earlier requested event starts strictly earlier. Detected activity is clipped to ten seconds. Event scores average within outputs, then within scenes, then across scenes, giving scenes equal weight.

\paragraph{Evaluation coverage.}
If any recognized word's alignment score is below 0.001, we omit its track from WER and speech timing. This excludes 32 speech-source evaluations (Table~\ref{tab:joint-coverage}). All 208 non-speech evaluations remain available: 78 generated stems and 130 SAM Audio extractions. WER covers all 15 scenes per method except Omni (14; one lacks accepted speech alignment); window success covers all 15 for every method. Appendix~\ref{app:evaluator-audit} examines alignment decisions and unavailable-speech effects.

\begin{table}[htbp]
    \centering
    \caption{Joint-evaluation coverage across generated outputs. Sources: scorable/requested. Matched events: detected/scorable requested events. WER scenes: scenes contributing to WER.}
    \label{tab:joint-coverage}
    \small
    \begin{tabular}{@{}lrrrr@{}}
        \toprule
        Method & Speech sources & Sound sources & Matched events & WER scenes\\
        \midrule
        Standard LTX-2.5 & 70/81 & 39/39 & 96/109 & 15\\
        Naive batch expansion & 72/81 & 39/39 & 109/111 & 15\\
        MiniMax H3 & 77/81 & 39/39 & 113/116 & 15\\
        Ovi & 76/81 & 39/39 & 106/115 & 15\\
        Gemini Omni & 25/27 & 13/13 & 38/38 & 14\\
        \methodname & 80/81 & 39/39 & 118/119 & 15\\
        \bottomrule
    \end{tabular}
\end{table}

\paragraph{Generation followed by source separation.}
To compare direct stem generation with single-track generation followed by source separation, we additionally use SAM Audio to extract each requested speech source from the single-track baselines. Together with the existing non-speech extractions, these stems are evaluated using the same metrics as \methodname's generated stems. Speech alignment is available for 248/270 extracted stems (91.9\%), and all 130 non-speech stems remain available. Requested utterances are detected in only 124/270 speech stems (45.9\%), with per-baseline coverage reported in Table~\ref{tab:sam-coverage}. Separation introduces an additional failure point before ASR and alignment because omitted or distorted speech can prevent matching the requested utterance. We therefore evaluate speech directly on the mixed audio in the original comparison. The reduced utterance detection also limits order evaluation, which requires both events in a pair to be detected, while missed events remain failures in window success.

\begin{table}[htbp]
    \centering
    \small
    \caption{Coverage after SAM Audio separation. Speech alignment and utterance detection are reported over all requested speech stems. Order coverage is scored pairs divided by non-overlapping requested pairs among available sources, including speech--sound pairs.}
    \label{tab:sam-coverage}
    \begin{tabular}{@{}lrrr@{}}
        \toprule
        Baseline & Speech alignment & Utterance detection & Order pairs \\
        \midrule
        LTX-2.5 & 74/81 & 17/81 & 2/62 \\
        Ovi     & 74/81 & 41/81 & 23/64 \\
        H3      & 74/81 & 51/81 & 36/62 \\
        Omni    & 26/27 & 15/27 & 7/21 \\
        \bottomrule
    \end{tabular}
\end{table}

\subsection{Source-Content Metrics}
\label{app:automatic-metrics}

\paragraph{Semantic source assignment.}
Using \texttt{laion/clap-htsat-unfused}, we compare each complete stem with every source description by cosine similarity of normalized audio/text embeddings. \emph{Target CLAP} is similarity to the intended description; \emph{competing CLAP} is the highest similarity to another source; their difference is the \emph{CLAP margin}. Assignment accuracy is the fraction of stems scoring highest for their intended description.

\paragraph{General-sound content.}
For non-speech sources, we compute CLAP against the source description in one-second segments at 0.5-second hops. A segment is in-window when at least half its duration lies inside the requested interval. We average scores separately inside and outside; their difference is the temporal CLAP margin, measuring stronger description agreement during the request. Full-clip sources use the complete recording.

\input{sections/appendix_human_evaluation}

\subsection{Validation and Sensitivity of the Automatic Evaluation}
\label{app:evaluator-audit}

Transcription, alignment, and extraction errors can distort activity-based scores; unavailable speech also changes which events enter averages. We audit these effects, event placement, containment tolerance, and scene-level uncertainty below.

\paragraph{Listening audit.}
For the original five-method comparison, one reviewer listened to 50 evaluated tracks and their original mixtures: all 20 alignment-rejected speech tracks, 15 accepted speech controls, and 15 non-speech tracks. They assessed transcript content, word timing, target presence, and other-sound activity. Rejected tracks represent 28 source exclusions across 19 outputs because speakers in a mixture share a transcript. Accepted controls comprise three per method: the lowest alignment score and two random tracks from different scenes.

Table~\ref{tab:evaluator-listening} gives the judgments. Of 20 rejected transcripts, 13 were judged correct (including one truncated ending recognized from context); four contained inaudible words, one omitted a repeated phrase, and two had substitutions. Of 15 accepted controls, 14 were correct and one uncertain, with untranscribed gibberish. Rejection thus affects both erroneous transcripts and intelligible speech.

Non-speech checks cover one cat, knock, and violin per method. Targets were audible in 10/15 evaluated tracks; four SAM Audio extractions omitted targets fully or partly audible in the mixture; one target was absent from both. Separately, activity was correct in seven tracks, non-target/noise in four, unassessable because no target was present in three, and uncertain in one. Non-target activity included \methodname's native Clockwork airship knock stem. In that scene, Omni's only order-scoring failure assigns identical starts to knocks and footsteps, consistent with extraction confusion. Extraction can remove targets, and residual sounds can masquerade as target activity; we interpret activity scores alongside CLAP assignment and human source-identity judgments.

\begin{table}[htbp]
    \centering
    \caption{Listening audit of the selected tracks; target presence and activity are assessed separately.}
    \label{tab:evaluator-listening}
    \small
    \begin{tabular}{@{}lr@{}}
        \toprule
        Judgment & Tracks\\
        \midrule
        Rejected speech: transcript judged correct & 13/20\\
        Rejected speech: inaudible words / omitted phrase / substitutions & 4 / 1 / 2\\
        Accepted speech: transcript judged correct & 14/15\\
        Accepted speech: uncertain, with untranscribed gibberish & 1/15\\
        \midrule
        Sound: target heard in evaluated track & 10/15\\
        Sound: target heard in mix but absent from extraction & 4/15\\
        Sound activity: correct / non-target / no target / uncertain & 7 / 4 / 3 / 1\\
        \bottomrule
    \end{tabular}
\end{table}

\paragraph{Unavailable speech evaluations.}
Alignment rejection can leave methods with different event sets. We recalculate saved results in two ways (Table~\ref{tab:audit-sensitivity}). First, counting rejected-track events as failures while holding other scores and averaging fixed lowers \methodname's window success from 93.1\% to 92.0\%; Omni stays at 85.0\%.

Second, we retain only sources scorable for all six methods and every evaluated seed in their scene; unavailability in any run excludes that source for all methods. Scorable sources with missing requested events remain failures. This common subset contains 22/40 sources and 352/640 method/seed event instances across 11 scenes, six with speech for WER. Window success is 89.4\% for \methodname, 75.0\% for Omni, and 34.3\% for MiniMax H3. \methodname ranks highest under both recalculations.

\begin{table}[htbp]
    \centering
    \caption{Unavailable-event sensitivity. The bound counts unavailable events as failures; common-subset columns retain only sources scorable for every method and seed.}
    \label{tab:audit-sensitivity}
    \small
    \begin{tabular}{@{}lrrrr@{}}
        \toprule
        Method & \makecell{Window-success\\bound $\uparrow$}
        & \makecell{Common\\window WER $\downarrow$}
        & \makecell{Common\\window success $\uparrow$}
        & \makecell{Common onset\\error (s) $\downarrow$}\\
        \midrule
        Standard LTX-2.5 & 18.7\% & 0.511 & 33.3\% & 1.20\\
        Naive batch expansion & 15.0\% & 0.730 & 26.5\% & 1.60\\
        Ovi & 23.5\% & 0.383 & 25.8\% & 2.01\\
        MiniMax H3 & 25.9\% & 0.348 & 34.3\% & 2.02\\
        Gemini Omni & 85.0\% & 0.111 & 75.0\% & 1.03\\
        \methodname & \textbf{92.0\%} & \textbf{0.000} & \textbf{89.4\%} & \textbf{0.29}\\
        \bottomrule
    \end{tabular}
\end{table}

\paragraph{Event placement and containment.}
Containment does not specify onset or interval occupancy. For matched events, onset error is the absolute difference between detected and requested starts; temporal intersection over union (IoU) is detected/requested interval overlap divided by their union. \methodname achieves 0.35~s and 0.663 versus Omni's 0.89~s and 0.469, indicating closer boundary alignment (Table~\ref{tab:joint-additional}). Containment remains primary because events may occupy only part of the request.

We separate explicitly timed sounds from less restrictive full-clip requests, with five scenes per category. Both \methodname and Omni reach 100.0\% for full-clip sounds, but 70.0\% versus 20.0\% for timed sounds, exposing a difference hidden by full-clip scores.

Requiring 100\% containment instead of 95\% removes the allowed 5\% outside activity, keeping detections, exclusions, and aggregation fixed. Strict success is 77.8\% for \methodname and 64.4\% for Omni; the carrier comparison gives 75.0\% with and 15.6\% without carriers (Table~\ref{tab:strict-window-success}). Both comparisons use joint speech/sound measurements over 15 scenes. Absolute scores fall, but \methodname remains highest and carriers retain their benefit.

\begin{table}[htbp]
    \centering
    \caption{Boundary diagnostics and window-success variants. Onset error and IoU use matched events. Timed and full-clip sound success each cover five scenes. Strict success requires 100\% containment; its main and carrier comparisons each cover 15 scenes.}
    \label{tab:joint-additional}
    \label{tab:strict-window-success}
    \small
    \setlength{\tabcolsep}{3pt}
    \begin{tabular}{@{}lrrrrr@{}}
        \toprule
        Method & \makecell{Onset\\error (s) $\downarrow$}
        & \makecell{Temporal\\IoU $\uparrow$}
        & \makecell{Timed sound\\success $\uparrow$}
        & \makecell{Full-clip sound\\success $\uparrow$}
        & \makecell{Strict window\\success $\uparrow$}\\
        \midrule
        Standard LTX-2.5 & 1.98 & 0.260 & 6.7\% & 83.3\% & 17.0\%\\
        Naive batch expansion & 2.98 & 0.173 & 3.3\% & 93.3\% & 14.4\%\\
        Ovi & 2.11 & 0.257 & 10.0\% & 76.7\% & 21.1\%\\
        MiniMax H3 & 1.71 & 0.327 & 3.3\% & 90.0\% & 20.9\%\\
        Gemini Omni & 0.89 & 0.469 & 20.0\% & \textbf{100.0\%} & 64.4\%\\
        \methodname & \textbf{0.35} & \textbf{0.663} & \textbf{70.0\%} & \textbf{100.0\%} & \textbf{77.8\%}\\
        \midrule
        \multicolumn{6}{@{}l}{Strict carrier comparison: with carriers \textbf{75.0\%}; without carriers 15.6\%.}\\
        \bottomrule
    \end{tabular}
\end{table}

\paragraph{Uncertainty across scenes.}
We average seeds within scenes and bootstrap paired scene differences with 5,000 draws (seed 20260911), including only scenes where both methods have the metric. For \methodname minus Omni, requested-window WER differs by $-0.008$ (95\% confidence interval (CI) $[-0.156,0.175]$; both means recomputed on the same 14 scenes), and window success by 8.1 percentage points (95\% CI $[-2.8,22.8]$; 15 scenes). Both intervals include zero, so direction is uncertain at this sample size. Onset error differs by $-0.541$~s (95\% CI $[-1.323,-0.068]$; 15 scenes), supporting lower error for \methodname. These intervals measure scene variation; Omni's single output per scene leaves its seed variation unmeasured.

%% file: sections/appendix_human_evaluation.tex
\subsection{Human Evaluation}
\label{app:human-evaluation}

The main study assesses naturalness, timing, and source separation; the component study assesses scene broadcast and entity routing beyond automatic content/activity scores.

\subsubsection{Presentation and Recorded Responses}

\paragraph{Consent and playback.}
Consent states that media are synthetic, participation is voluntary, anonymized judgments are aggregated, and participants may stop anytime. We collect no demographics. We recommend headphones, a quiet room, and a desktop/laptop. Unscored tutorials introduce playback and response controls; replay is unrestricted and audio uses its stored level.

\paragraph{Blinding and randomization.}
Neutral labels conceal methods; scene order and candidate labels are shuffled independently per submission. Component trials present naturalness before routing, shuffling scenes within blocks. Browser progress persists until submission; records contain judgments, presentation order, trial durations, play counts, and playback time.

\paragraph{Response counts.}
The main study has \MainResponses{} completed responses over 15 scenes: \MainTrials{} rankings per method per ranking task and \MainTrials{} separation choices. The component study has \AblationResponses{} completed responses, each with ten naturalness and ten routing trials: \AblationTrials{} judgments per condition and measure. Naturalness covers 14 scenes across submissions, ten per response.

\subsubsection{Main Comparison: Naturalness, Timing, and Separation}

Trials show Standard LTX-2.5, naive batch expansion, and \methodname with a scene brief and source intervals. Listeners:
\begin{enumerate}
    \item Rank \textbf{overall naturalness}: sound clarity/coherence, natural voices/emotion, and whether sounds belong together.
    \item Rank \textbf{timeline adherence} to requested windows using timelines, waveforms, and playback.
    \item Compare mixes and isolated tracks from the two multi-stem methods and choose the cleaner \textbf{source separation}; \emph{Both} records a tie.
\end{enumerate}
Rankings use distinct ranks 1 (best) to 3; separation ties give each method half-credit. Stimuli comprise 45 audiovisual candidates, 30 multi-stem sets, and 80 isolated stems (40 requested sources per multi-stem method).

\subsubsection{Component Comparison: Scene Broadcast and Entity Routing}

\paragraph{Contextual naturalness and sound quality.}
Naturalness scenes (Table~\ref{tab:additional-scenes}) test whether broadcast makes two separately generated speech stems fit a shared interaction. Visual prompts specify interaction/delivery; source prompts specify voices/words. Intended deliveries accompany candidates; Hospital farewell is the unscored tutorial.

We compare Stem Formation audio paired with full-method video, Scene Integration without broadcast, and the full method; both integration conditions use their respective audiovisual outputs and share initial Stem Formation states, masks, timeline, and refinement seed.

Listeners rank contextual naturalness (emotion, interaction, volume, tone, texture, shared environment) and overall listening preference, and score each candidate's sound quality 0--10: unusable/heavily distorted (0), understandable with noticeable problems (5), clean/natural (10).

\paragraph{Visual activation and ownership.}
Routing trials use nine benchmark scenes---Noir office, Enchanted archive, Orbital greenhouse, Android laboratory, Astral conservatory, Neon biology laboratory, Clockwork airship, Podcast studio, and Studio introduction (Table~\ref{tab:benchmark-scenes})---plus Market conversation; Studio conversation is the unscored tutorial (Table~\ref{tab:additional-scenes}). Routing on/off yields 20 scored audiovisual candidates. For each, listeners answer Yes/No: do all intended visible sources react at the correct times (\emph{target activation}), and do only those sources react, without unrelated entities responding (\emph{correct ownership})?

\subsubsection{Aggregation}

We pool completed-response judgments by condition, reporting mean ranks, mean sound quality, affirmative-response percentages, and response counts. First-place frequency is the rank-one fraction; separation preferences use half-credit for ties.

%% file: sections/appendix_additional_results.tex
\section{Additional Results and Ablations}
\label{app:additional-results}

Evaluation protocols are in Appendix~\ref{app:evaluation-protocols}.

\subsection{Generation Comparison}

We assess intended source content with CLAP and audiovisual coherence through listening judgments of separation, naturalness, and timing. Source assignment is the fraction of stems whose intended description has the highest scene-source CLAP score; non-speech CLAP measures agreement with requested sounds. After averaging seeds within each scene, we form 95\% confidence intervals (CIs) from 2,000 bootstrap resamples of evaluated scenes, drawing scenes with replacement and recomputing the mean (seed 2027). The same evaluation extends to 30 additional scenes, using 20,000 bootstrap resamples for more stable intervals.

\methodname achieves 90.0\% source assignment versus naive expansion's 82.2\%: a paired gain of 7.8 percentage points (95\% CI $[-6.7,20.6]$). Non-speech CLAP gains are 0.093 over standard LTX-2.5 (95\% CI $[0.057,0.129]$) and 0.098 over naive expansion (95\% CI $[0.029,0.169]$), supporting closer sound-description agreement. The source-assignment point estimate favors \methodname, but its interval includes zero. With the additional scenes, assignment is 91.6\% versus 82.9\%, a paired gain of 8.7 points (95\% CI $[2.7,14.4]$), supporting higher accuracy. Detailed configurations for the expanded evaluation are included in the supplementary material.

We further compare \methodname's directly generated stems with single-track generation followed by SAM Audio separation on the original 15 scenes (Table~\ref{tab:sam-all-stems}). \methodname achieves lower requested-window WER (0.060 versus 0.444--0.872), higher window success (93.1\% versus 15.0--57.2\%), and less off-window activity (1.7\% versus 40.8--59.3\%). Source-assignment accuracy is also higher (90.0\% versus 50.0--55.0\%), indicating that the generated stems more consistently match their intended source descriptions. These results favor direct generation with source-level control over post-hoc separation for both requested timing and source content. Appendix~\ref{app:joint-audio-metrics} describes the separation procedure and evaluation coverage.

\begin{table}[htbp]
    \centering
    \small
    \caption{Direct stem generation versus single-track generation followed by SAM Audio separation on 15 scenes. Window success uses 95\% activity containment. Full-clip sound success covers five scenes with unrestricted sound timing. Order accuracy is conditional on detected event pairs; Table~\ref{tab:sam-coverage} reports coverage for the separated baselines.}
    \label{tab:sam-all-stems}
    \resizebox{\linewidth}{!}{%
        \begin{tabular}{@{}lrrrrrr@{}}
            \toprule
            Method & \makecell{Window\\WER $\downarrow$}
            & \makecell{Window\\success $\uparrow$}
            & \makecell{Full-clip sound\\success $\uparrow$}
            & \makecell{Order\\accuracy $\uparrow$}
            & \makecell{Off-window\\activity $\downarrow$}
            & \makecell{Source\\assignment $\uparrow$}\\
            \midrule
            LTX-2.5 + SAM Audio & 0.872 & 15.0\% & 83.3\% & \textbf{100.0\%} & 53.7\% & 50.6\%\\
            Ovi + SAM Audio & 0.724 & 19.8\% & 76.7\% & 96.8\% & 59.3\% & 50.0\%\\
            H3 + SAM Audio & 0.563 & 22.2\% & 90.0\% & 92.2\% & 53.4\% & 52.2\%\\
            Omni + SAM Audio & 0.444 & 57.2\% & \textbf{100.0\%} & 73.3\% & 40.8\% & 55.0\%\\
            \midrule
            \methodname & \textbf{0.060} & \textbf{93.1\%} & \textbf{100.0\%}
            & \textbf{100.0\%} & \textbf{1.7\%} & \textbf{90.0\%}\\
            \bottomrule
        \end{tabular}%
    }
\end{table}

Of \MainTrials{} main-study separation choices, \MainSepFullWins{} favor \methodname, \MainSepNaiveWins{} favor naive expansion, and \MainSepTies{} are ties. Half-credit per tie yields \MainFullSep{}\% versus \MainNaiveSep{}\% preference (Table~\ref{tab:human-comparison}). \methodname receives \MainNatWins{} first-place audiovisual-naturalness rankings (\MainFullNatFirst{}\%) and \MainTimeWins{} first-place timing rankings (\MainFullTimeFirst{}\%), favoring both source separation and the combined audiovisual output.

\begin{figure}[!t]
    \centering
    \includegraphics[width=\linewidth,height=\textheight,keepaspectratio,clip]{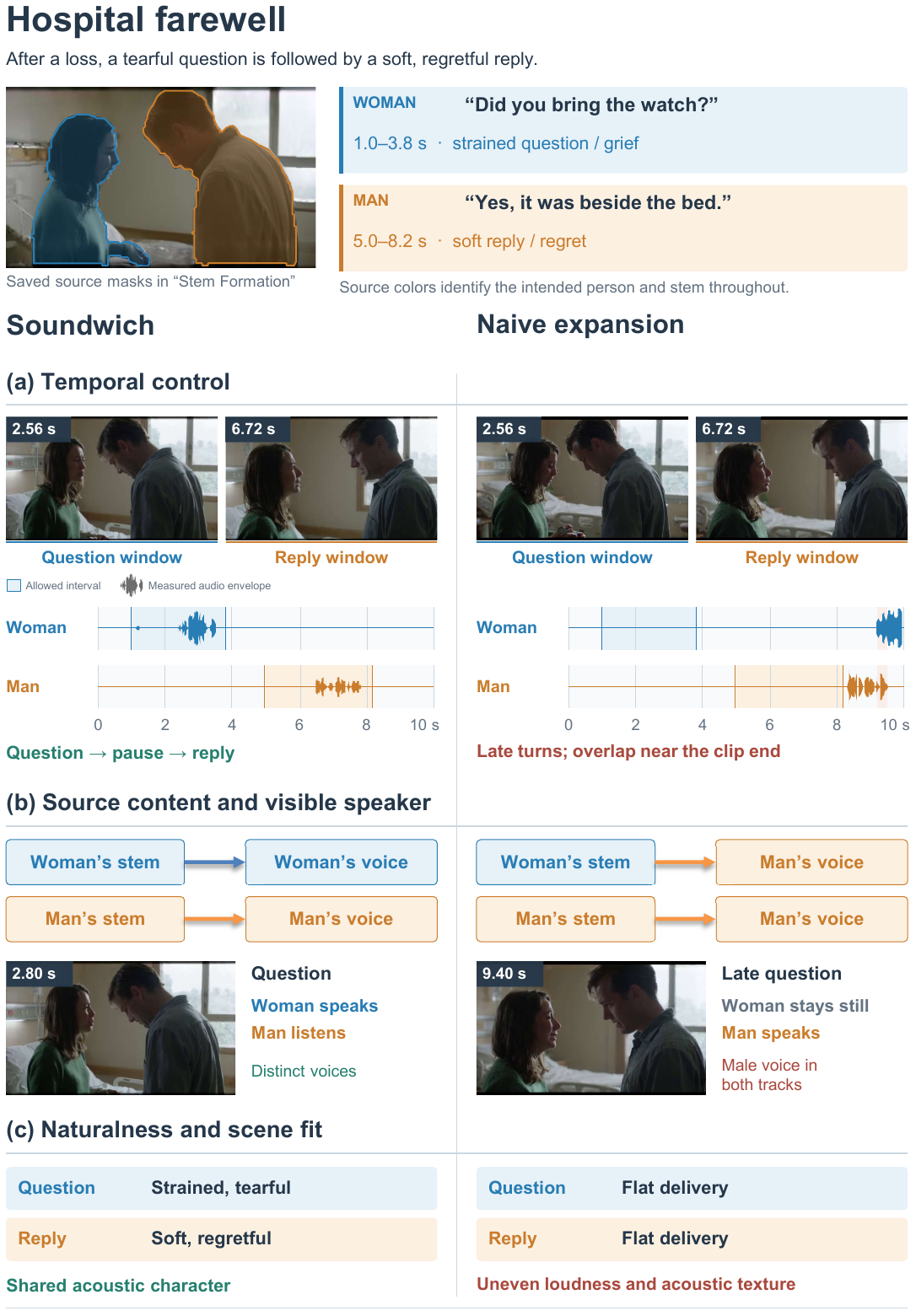}
    \caption{\textbf{Hospital farewell.} Comparison of \methodname and naive batch expansion: \textbf{(a)} temporal control, \textbf{(b)} source content and visible speaker, and \textbf{(c)} naturalness and scene fit. Source colors link the visible entities to their audio stems, and shaded intervals indicate the requested speaking windows. 
    }
    \label{fig:hospital-results}
\end{figure}

Figure~\ref{fig:hospital-results} compares timing, voice separation, visible speaking activity, and naturalness in Hospital farewell, the empty-hospital-bed conversation discussed in Section~\ref{sec:generation-comparison}.

\subsection{Quiet Carrier versus Zero Suppression Target}
\label{app:zero-vs-quiet}

We test whether suppressing features toward zero suffices to silence a source outside its requested intervals, versus using a quiet-reference carrier. In 15 paired scenes, the zero target replaces the quiet carrier in both Stem Formation and Scene Integration; seeds, suppression strengths, and other settings are fixed within pairs. Each variant derives masks from its own Stem Formation video. We assess refined stems by listening, in-/off-window waveform power, speech WER, and non-speech CLAP.

Zero-target outputs contain audible random noise during requested silence; quiet carriers keep these intervals quiet. Without quiet-reference features, the model may generate sounds from its learned prior. Quiet carriers reduce off-window power, retain more in-window power, lower WER, and raise non-speech CLAP, preserving requested content (Table~\ref{tab:zero-vs-quiet}).

\begin{table}[htbp]
    \centering
    \caption{Suppression targets in 15 paired scenes, one seed each. Power is mean squared waveform amplitude, averaged over controlled stems within scenes, then scenes; off-window power excludes 0.20-second boundary transitions. Full-window WER uses Whisper word timestamps, including every recognized word with any overlap with the requested interval. Non-speech CLAP covers eight eligible scenes.}
    \label{tab:zero-vs-quiet}
    \small
    \begin{tabular}{@{}lrrr@{}}
        \toprule
        Metric & Quiet carrier & Zero target & Scenes\\
        \midrule
        Off-window power $\downarrow$ & $\mathbf{1.61\times10^{-7}}$ & $1.52\times10^{-3}$ & 15\\
        In-window power & $1.66\times10^{-2}$ & $9.62\times10^{-3}$ & 15\\
        Full-window WER $\downarrow$ & \textbf{0.120} & 0.600 & 15\\
        Non-speech CLAP $\uparrow$ & \textbf{0.431} & 0.240 & 8\\
        \bottomrule
    \end{tabular}
\end{table}

\subsection{Activation Placement within an Allowed Window}
\label{app:activation-direction}

Qualitative Gemini Omni comparisons showed repeated utterances filling requested windows. Our default weak, uniform activation leaves onset, duration, pauses, and silence to the model. We test uniform activation against linear early emphasis (peak to zero) and late emphasis (zero to peak). All share peak strength; early/late have equal total activation, while uniform has twice as much.

Three single-speaker scenes (Quiet room, Reception desk, Quiet garden; Table~\ref{tab:additional-scenes}) use $[0.5,9.5]$-second windows and three paired seeds each, yielding 27 outputs. Profiles weight activation after RMS normalization. The scene stem is disabled; suppression, boundary transitions, and denoising settings are fixed. All profiles share a brief initial denoising boost, then lower strength. Whisper transcription and wav2vec2 word alignment measure first-sentence onset and complete-sentence repetitions in Stem Formation audio.

Late emphasis delays mean first onset in every scene, by 4.018~s on average relative to early emphasis (Table~\ref{tab:activation-direction}). Repetition occurs in 3/9 uniform, 2/9 early, and 0/9 late outputs. Some uniform Quiet room/garden outputs contain early and late utterances; late emphasis removes the early occurrence while retaining the late one. Another garden sample develops a long within-sentence pause. Profiles thus affect both timing and which occurrences survive: late emphasis can remove early repetitions but biases speech toward the window's end. We retain uniform activation to avoid an early/late preference.

\begin{table}[htbp]
    \centering
    \caption{Mean first-sentence onset (seconds), three paired seeds per scene. U/E/L: uniform/early/late activation; L$-$E is the paired difference. The mean weights scenes equally.}
    \label{tab:activation-direction}
    \small
    \begin{tabular}{@{}lrrrr@{}}
        \toprule
        Scene & U & E & L & L$-$E\\
        \midrule
        Quiet room & 1.697 & 1.697 & 8.024 & +6.327\\
        Reception desk & 6.314 & 6.514 & 7.676 & +1.162\\
        Quiet garden & 4.283 & 1.844 & 6.407 & +4.563\\
        \midrule
        Mean & 4.098 & 3.352 & 7.369 & +4.018\\
        \bottomrule
    \end{tabular}
\end{table}

\subsection{Source-specific Guidance for Source Separation}
\label{app:negative-cfg-diagnostic}

Source-specific classifier-free guidance (CFG) discourages sounds assigned to other stems. We test leakage suppression and target preservation in six scenes of associated concurrent sounds (Table~\ref{tab:additional-scenes}), screening paired outputs for clearly audible leakage. Pairs share scene and seed; carriers, timeline control, and scene stem are disabled. We listen to Stem Formation stems and measure CLAP against intended/competing descriptions, averaging stems within scenes and weighting the six scenes equally.

Without source-specific CFG, Theater curtain's orchestra stem contains applause; guidance reduces it while retaining recognizable orchestra. Similar intrusions involve saxophone and other Blues quartet instruments, and jackhammer noise and other Construction site sounds. Listening identifies eight intrusions across six scenes; CLAP similarity to the intruder falls in all eight paired comparisons. Scene-level competing CLAP falls, while target CLAP and assignment improve (Table~\ref{tab:negative-cfg-diagnostic}). These leakage cases support suppressing competing sounds when leakage occurs.

\begin{table}[htbp]
    \centering
    \caption{Source-specific guidance in six paired leakage cases, evaluating the same Stem Formation stems in both variants. Competing CLAP is the maximum similarity to another scene-source description; margin is target minus competing CLAP.}
    \label{tab:negative-cfg-diagnostic}
    \small
    \begin{tabular}{@{}lrrrr@{}}
        \toprule
        Variant & \makecell{Target CLAP $\uparrow$} & \makecell{Competing CLAP $\downarrow$} & \makecell{CLAP margin $\uparrow$} & \makecell{Source\\assignment $\uparrow$}\\
        \midrule
        With source-specific guidance & \textbf{0.344} & \textbf{0.115} & \textbf{0.229} & \textbf{95.8\%}\\
        Without source-specific guidance & 0.313 & 0.262 & 0.051 & 61.1\%\\
        \bottomrule
    \end{tabular}
\end{table}

\subsection{Scene Broadcast Results}

We test whether scene broadcast during Scene Integration improves conversational delivery and shared acoustic character. The Contextual naturalness scenes in Table~\ref{tab:additional-scenes} compare broadcast on/off from identical Stem Formation states, masks, timelines, and refinement seeds. Stem Formation audio, paired with full-method video, provides a refinement reference.

Listeners rank contextual naturalness (emotion, interaction, volume, tone, shared environment) and overall listening preference, and rate sound quality 0--10. There are \AblationTrials{} judgments per condition from \AblationResponses{} completed responses; Appendix~\ref{app:human-evaluation} gives presentation and aggregation details.

Broadcast improves mean context rank from \NoBroadcastContext{} to \FullContext{} (lower is better), overall preference, and sound quality (Table~\ref{tab:broadcast-full-results}), supporting more coherent interaction between separately generated voices. Both integration variants improve quality over Stem Formation audio; broadcast adds a further gain.

\begin{table}[htbp]
    \centering
    \caption{Scene-broadcast results, including Stem Formation audio. Each condition has \AblationTrials{} judgments from \AblationResponses{} completed responses. Context/preference: mean ranks 1 (best)--3; quality: 0--10.}
    \label{tab:broadcast-full-results}
    \small
    \begin{tabular}{@{}lrrr@{}}
        \toprule
        Measure & \makecell{%
Stem Formation\\audio} & \makecell{%
Scene Integration\\without broadcast} & \makecell{%
Scene Integration\\with broadcast}\\
        \midrule
        Context rank $\downarrow$ & \StageOneContext{} & \NoBroadcastContext{} & \textbf{\FullContext{}}\\
        Quality score $\uparrow$ & \StageOneQuality{} & \NoBroadcastQuality{} & \textbf{\FullQuality{}}\\
        Preference rank $\downarrow$ & \StageOneOverall{} & \NoBroadcastOverall{} & \textbf{\FullOverall{}}\\
        \bottomrule
    \end{tabular}
\end{table}

\subsection{Direct Sharing between Source Stems}
\label{app:direct-sharing}

We test whether direct cross-stem self-attention preserves source content, comparing it with independent attention. Our method exchanges context through an intermediate scene stem. Three scenes contain two speakers, instruments, or percussion sources (Table~\ref{tab:additional-scenes}); on/off pairs share scene and seed, with independent noise per stem. Carriers, timeline control, scene stem, and routing are disabled.

Sharing concatenates both stems' audio tokens for full-strength joint self-attention at every block and denoising step in both Stem Formation and Scene Integration, preserving original temporal positions. Without sharing, self-attention stays independent. We assess isolated Stem Formation stems by listening, complete speech transcripts, and CLAP audio-embedding cosine similarity.

Direct sharing makes stems sound nearly identical, losing distinct requested content. Without sharing, speech transcripts are ``The garden smells of roses'' and ``My train leaves at 7 o'clock.'' With sharing, both read ``My garden leaves smells of roses,'' mixing both utterances. Similarity rises in all three scenes, by 0.458 on average (Table~\ref{tab:direct-sharing}).

Untwisting RoPE~\cite{mikaeili2026untwisting} describes image copying driven by high-frequency rotary-position components favoring reference tokens at matching spatial positions. Shared audio temporal coordinates may similarly favor copying at matching times. Our gather/broadcast uses this effect: Stem Formation copies source structure into the scene stem; Scene Integration copies scene structure back, retaining separate output-stem self-attention.

\begin{table}[htbp]
    \centering
    \caption{CLAP audio-embedding cosine similarity between two isolated Stem Formation stems, one paired seed per scene. Higher values mean more similar embeddings.}
    \label{tab:direct-sharing}
    \small
    \begin{tabular}{@{}lrrr@{}}
        \toprule
        Scene & Sharing off & Sharing on & Difference\\
        \midrule
        Speakers & 0.640 & 0.972 & +0.332\\
        Musicians & 0.336 & 0.973 & +0.637\\
        Percussion & 0.558 & 0.963 & +0.406\\
        \bottomrule
    \end{tabular}
\end{table}

\subsection{Entity Routing Evaluation}
\label{app:routing-evaluation}

Across ten scenes, listeners compare paired routing-on/off outputs for \emph{target activation} (intended visible sources react at requested times) and \emph{correct ownership} (only those sources react). \AblationResponses{} completed responses yield \AblationTrials{} judgments per variant and question; Appendix~\ref{app:human-evaluation} gives the protocol.

Routing improves affirmative judgments for both measures (Table~\ref{tab:routing-human-evaluation}), supporting correctly timed sound-to-entity association. In Studio introduction, a woman and man speak in separate turns; all responses affirm routed ownership, versus \StudioUnroutedYes{} of \AblationResponses{} without routing.

\begin{table}[htbp]
    \centering
    \caption{Entity routing across ten scenes: affirmative-response rates from \AblationTrials{} judgments per variant and question.}
    \label{tab:routing-human-evaluation}
    \small
    \begin{tabular}{@{}lrr@{}}
        \toprule
        Entity routing & Target activation $\uparrow$ & Correct ownership $\uparrow$\\
        \midrule
        With routing & \textbf{\FullActivation{}\%} & \textbf{\FullOwnership{}\%}\\
        Without routing & \NoRoutingActivation{}\% & \NoRoutingOwnership{}\%\\
        \bottomrule
    \end{tabular}
\end{table}

\suppressfloats[t]

%% file: sections/appendix_applications.tex
\section{Generalization and Application}
\label{app:editing-efficiency-limitations}

\subsection{Cross-Backbone Transfer to Ovi}
\label{sec:portability-generalization-results}

We transfer batch expansion, source-specific guidance, and carrier timeline control to Ovi~\cite{low2025ovi}, retaining source descriptions, competing sounds to exclude, and activity intervals. Ovi-generated carrier features are replayed at its audio self-attention outputs.

We use the Ovi~1.1 \texttt{960x960\_10s} checkpoint in text-to-video mode: 10-second clips at $1280\times704$, 50 UniPC steps, timestep shift $5.0$, audio/video guidance $3.0$/$4.0$, and skip-layer guidance at layer~11, with INT8 quantization and CPU offloading.

The activation reference depicts a loud party with music and cheering; suppression requests silence over a black frame. Both use seed $42$ and the target sampling settings. Effective activation strength is $0.12$ and quiet-carrier blending weight $0.50$; carrier token selection and boundary transitions follow Appendix~\ref{app:cached-carriers}.

Table~\ref{tab:additional-scenes} details conversations with two or three speech sources: a passenger requests tickets, a clerk gives departure information, and the passenger responds (ticket counter, seed $3$); teacher and children alternate (classroom, seed $36$); three speakers take successive turns (auction, seed $101$). With all three components active, Ovi produced separate speech stems and could reproduce requested utterances in order, demonstrating independently conditioned audio and carrier scheduling on a second joint video-audio backbone.

We also attempted entity-routed refinement by re-noising the initial video to align speech with visible speakers. Routing did not reliably eliminate wrong-speaker motion. Stronger refinement could misalign fixed masks by changing the scene; increased A2V attention gain introduced visual artifacts. The three audio controls transferred successfully, but reliable visual correspondence remained unresolved.

\subsection{Cross-Backbone Transfer to MiniMax H3}
\label{app:h3-transfer}

We also transfer \methodname{} to MiniMax H3~\cite{minimax2026h3}, extending source-separated generation, timeline control, entity-routed refinement, and editing to a frozen single-stream backbone. H3 jointly attends over text, video, and audio tokens. Shared video features convey context across sources, supporting contextually appropriate delivery and broadly consistent acoustic character in our qualitative inspection; we therefore omit the additional scene stem.

For $N$ sources, we construct $N+1$ batch rows. Before each transformer block, shared video features are copied into the source rows, while the video row attends to concatenated audio keys and values from all sources. Each audio row receives an isolated source description within neutral scene context. H3-generated carrier features are replayed at the audio-token outputs of joint attention, and entity routing operates within this joint sequence.

We apply the transfer to the full benchmark and demonstrate both retiming and content replacement: moving the cat's vocalization in Enchanted Archive and replacing the detective's line in Occult Noir. Video refinement is conditioned on the edited audio latents, which remain fixed.

Clips contain 243 frames at 24~fps and $768\times448$, with 23 model evaluations per stage. Refinement uses $\sigma_v=\sigma_a=0.95$, quiet-reference blending weight $0.3$, and outside-window attenuation $0.25$; editing retains the same video noise level with clean, fixed audio.

\subsection{Independent Timeline and Stem Editing}
\label{app:stem-editing}

Independent stems allow us to change the timing or content of a sound while retaining the other generated audio. Editing first constructs the revised audio latents, and then reruns video refinement with those latents held fixed.

\paragraph{Editing the stems.}
Retiming moves the utterance portion of each saved audio latent and rearranges surrounding tokens to preserve clip length. For example, reversing two speakers' order moves the first line later and the second earlier, reusing both utterances; revised times guide video refinement.

Content replacement runs Stem Formation for one output stem with the revised source prompt, original visual prompt, and original clip duration. Its audio latent replaces the selected stem; others retain their saved Scene Integration latents. The replacement can also be retimed.

\paragraph{Adapting the video.}
We restart from the saved Stem Formation video and its entity masks, spatially upscaling and re-noising with the Scene Integration initialization and eight-step schedule (Appendix~\ref{app:generation-settings}). Only video latents are updated; edited audio latents remain clean and fixed. Routing uses revised timelines and original spatial masks. The scene stem is disabled; suppression replay remains active outside revised intervals. The refined video is paired with decoded edited stems. Table~\ref{tab:editing-stages} compares generation and editing controls.

\begin{table}[!ht]
    \centering
    \caption{Generation and editing controls. During editing refinement, audio is fixed and entity routing conditions video on the revised stems.}
    \label{tab:editing-stages}
    \small
    \setlength{\tabcolsep}{5pt}
    \begin{tabular}{@{}lccc@{}}
    \toprule
    Operation &
Stem Formation &
Scene Integration & Editing refinement \\
    \midrule
    Source audio updated & Yes & Yes & No \\
    Activation replay & Yes & No & No \\
    Suppression replay & Yes & Yes & Yes \\
    Scene context & Gather & Broadcast & Disabled \\
    Entity routing & No & Yes & Yes \\
    \bottomrule
    \end{tabular}
\end{table}

\subsection{Efficiency}
\label{sec:runtime_and_memory}

Each source adds an audio branch while sharing video computation. Profiling uses one 32~GB NVIDIA RTX~5090 and 249-frame clips at 25~fps: Stem Formation runs 40 denoising steps at $384\times224$; Scene Integration refines spatially by $2\times$ to $768\times448$. Two or more output sources include an internal scene stem.

Timing includes pipeline construction, inference, and artifact writing for both stages. Cached carriers and deterministic disjoint routing masks exclude carrier construction, SAM segmentation, and manual mask correction from timing. Each stem count has three runs with fresh processes for each stage. We report medians of total runtime (summed stage times) and peak allocated memory (the larger stage-wise peak PyTorch tensor allocation).

\begin{figure}[htbp]
    \centering
    \includegraphics[width=\linewidth]{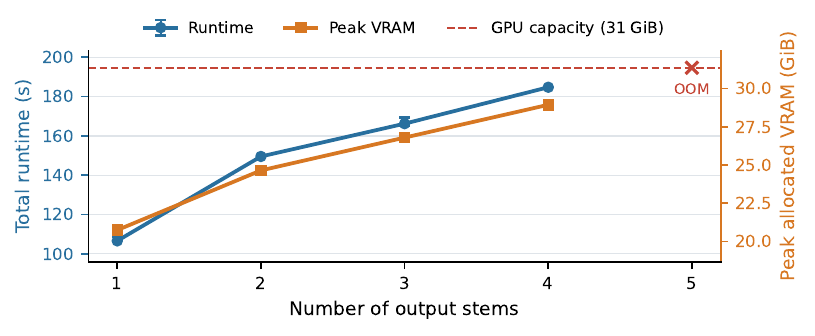}
    \caption{Runtime and peak allocated GPU memory versus output-stem count on one NVIDIA RTX~5090: medians of three repeats with cached carriers and deterministic routing masks. Two or more sources enable the scene stem; five exceed memory in Scene Integration.}
    \label{fig:stem-scaling}
\end{figure}

From one to four output stems, runtime rises from 106.6 to 184.7 seconds and peak allocation rises from 20.7 to 28.9~GiB; both grow more slowly than source count (Figure~\ref{fig:stem-scaling}). The largest runtime jump is from one to two sources, when the scene stem is introduced. Five stems exceed memory in Scene Integration, limiting this configuration to four.

\subsection{Observed Failure Modes}

\paragraph{Timing and source activation.}
Sounds may be weak or absent depending on the seed and activation-carrier match. Stronger activation can help, but may cause robotic speech or audible carrier artifacts. Very short windows or complex activity sequences reduce timing reliability and can cause omissions or incorrect event order.

\paragraph{Shared context.}
With overlapping sounds, persistent sources such as music can dominate scene-stem features, reducing shorter events' contribution to broadcast context. Excluding the dominant source from gathering and broadcast lets the scene stem focus on other sounds; the excluded source's independent stem remains in the final mix.

\paragraph{Spatial ownership and revision.}
Small or occluded entities may have incomplete masks, leaving visible regions without intended audio conditioning. Higher A2V attention gain can strengthen responses but cause local color or motion discontinuities. Editing refinement may also change appearance or motion beyond the revised event, giving local audio edits broader visual consequences.